\documentclass[sigconf]{acmart}
\AtBeginDocument{%
  \providecommand\BibTeX{{%
    \normalfont B\kern-0.5em{\scshape i\kern-0.25em b}\kern-0.8em\TeX}}}
\usepackage{tabularray}
\UseTblrLibrary{siunitx, varwidth}
\usepackage{etoolbox}
\usepackage{amsmath}
\usepackage{multirow} 
\usepackage{soul}
\renewcommand\footnotetextcopyrightpermission[1]{} 
\usepackage{graphicx}
\usepackage{subcaption}
\usepackage{xcolor}
\usepackage{mwe}
\usepackage{xspace}
\usepackage{algorithm}
\usepackage{algorithm,algpseudocode}
\usepackage{gensymb}
\newcommand{\ie}{\emph{i.e.,}\xspace}
\newcommand{\eg}{\emph{e.g.,}\xspace}

\newcommand{\G}{{\mathcal G}}

\newcommand{\eat}[1]{}
\newcommand{\warn}[1]{{\color{red}{#1}}}

\newcommand{\sstab}{\rule{0pt}{8pt}\\[-2.4ex]}

\newcommand{\bi}{\begin{itemize}}
\newcommand{\ei}{\end{itemize}}
        {\end{itemize}\vspace{-1.5ex}}

\newcommand{\stitle}[1]{\vspace{.5ex}\noindent{\bf #1}}
\newcommand{\eetitle}[1]{\vspace{0.8ex}\noindent{\underline{\em #1}}}

\definecolor{gray}{rgb}{0.5,0.5,0.5}

\newcommand{\stdyn}{ST-GTrend\xspace}

\newcommand{\parastdyn}{Para-GTrend\xspace}

\begin{document}
\setcopyright{none}

% \title{Photovoltaic Degradation Analysis at Scale \\ using
% Parallel-friendly Spatio-Temporal Graph Learning}
\title{Parallel-friendly Spatio-Temporal Graph Learning for Photovoltaic Degradation Analysis at Scale}
% \title{Spatio-Temporal Dynamic Graph Neural Networks \\
%  for Long-term Trend Analysis in Complex Physical Systems}

% \title{Analyzing Degradation Patterns in Complex Physical Systems \\ with Spatio-Temporal Dynamic Graph Neural Network}

% \title{Spatio-Temporal Dynamic Graph Neural Network \\
% for Long-term Trend Analysis}

% \author{Yangxin Fan, Raymond Wieser}
% \affiliation{%
%   \institution{Case Western Reserve University
% }
% %\country{USA}
% }
% \email{{yxf451, rxw497}@case.edu}

% \author{Laura S.Bruckman, Roger H.French, Yinghui Wu}
% \affiliation{%
%   \institution{Case Western Reserve University
% }
% %\country{USA}
% }
% \email{{lsh41,rxf131,yxw1650}@case.edu}

\author{Yangxin Fan, Raymond Wieser, Laura Bruckman, Roger French, Yinghui Wu}
\affiliation{%
  \institution{Case Western Reserve University}
  \city{Cleveland}
  \state{Ohio}
  \country{USA}}
% \affiliation{%
%   \institution{Case Western Reserve University, Cleveland, Ohio, USA}}
\email{{yxf451, rxw497, lsh41, rxf131, yxw1650}@case.edu}

% \author{Laura S.Bruckman, Roger H.French, Yinghui Wu}
% \affiliation{%
%   \institution{Case Western Reserve University
% }
% %\country{USA}
% }
% \email{{lsh41,rxf131,yxw1650}@case.edu}

\begin{abstract}
We propose a novel Spatio-Temporal Graph Neural Network empowered trend analysis approach (\stdyn) to perform fleet-level performance degradation analysis for 
Photovoltaic (PV) power networks. PV power stations have become an integral component to the global sustainable energy production landscape. 
Accurately estimating the performance of PV systems is critical to their feasibility as a power generation technology and as a financial asset.
One of the most challenging problems in assessing the Levelized Cost of Energy (LCOE) of a PV system is to understand and estimate the long-term Performance Loss Rate (PLR) for large fleets of PV inverters. 
% new abbreviation: G-Trend, st-Trend-GNN
\stdyn integrates spatio-temporal coherence and graph attention 
to separate PLR 
as a long-term ``aging'' trend from 
multiple fluctuation terms 
in the PV input data. 
To cope with diverse degradation patterns in timeseries, \stdyn adopts a paralleled graph autoencoder array to extract aging and fluctuation terms simultaneously. 
\stdyn imposes flatness and smoothness regularization to ensure the disentanglement between aging and fluctuation. To scale the analysis to 
large PV systems, we also introduce \parastdyn, a %hybrid (model and data) 
parallel algorithm to accelerate 
the training and inference of \stdyn.
We have evaluated \stdyn on three large-scale PV datasets, spanning a time period of 10 years. 
Our results show that \stdyn reduces Mean Absolute Percent Error (MAPE) and Euclidean Distances by 34.74\% and 33.66\% compared to the SOTA methods.
Our results demonstrate that \parastdyn can speed up \stdyn by up to 7.92 times. 
We further verify the generality and 
effectiveness of 
\stdyn for trend analysis using  financial and economic datasets. 
Our source code, datasets, and a full version of the paper are made available. \footnote{\url{https://github.com/Yangxin666/ST-GTrend}}
\end{abstract}

\maketitle

\section{Introduction}
\label{sec:introduction}
Photovoltaic energy represents a promising solution to growing uncertainty over the stability of the world's energy resources and lowering the carbon footprint~\cite{gazbour2018path}. 
For large-scale deployment of this well established technology, the financial benefits of investing in commodity scale solar must be highlighted to encourage the adoption of PV power generation. 
This is best accomplished through careful accounting of all of the possible costs associated with the production of energy.
The Levelized Cost of Energy (LCOE)~\cite{darling2011assumptions} is a measurement of the total cost of the generation of one unit of energy, and is used to quantify the profitability of a technology. 
The LCOE summarizes the lifetime of the system including, initial cost for planning, building, maintenance, and destruction of the site. 
A significant portion of the LCOE of a PV system is related to its ability to produce the same amount of power under the same weather conditions ({\em Performance Loss}) \cite{jordanPVDegradationCurves2017}. 
Performance loss in PV modules on an individual's roof may impact a single family electricity bill, but the performance loss across fleets PV power plants contribute to millions of dollars in lost revenue. 
The {\em Performance Loss Rate} (PLR) of a photovoltaic (PV) system indicates the decline of the power output over time (in units of \% per annum (\%/a, or \%/year)) \cite{french2021assessment}. 
PLR represents the physical degradation of PV modules, which is a %unavoidable 
critical property of any PV system. 
There are two approximating models to 
quantify PLR: relative PLR \cite{lindig2018review}, aggregated by comparing power data output to its counterpart from initial state of the PV system and absolute PLR \cite{ingenhoven2017comparison}, which is the coefficient of the slope 
of the power output as time goes by. 
\eat{Both assume a linear degradation rate, which cannot characterize diversified degradation patterns in real-world PV systems \cite{jordanPVDegradationCurves2017}.}

 \begin{figure}[tb!]
  \vspace{-2ex}
  \centering
  \includegraphics[width=0.95\linewidth]{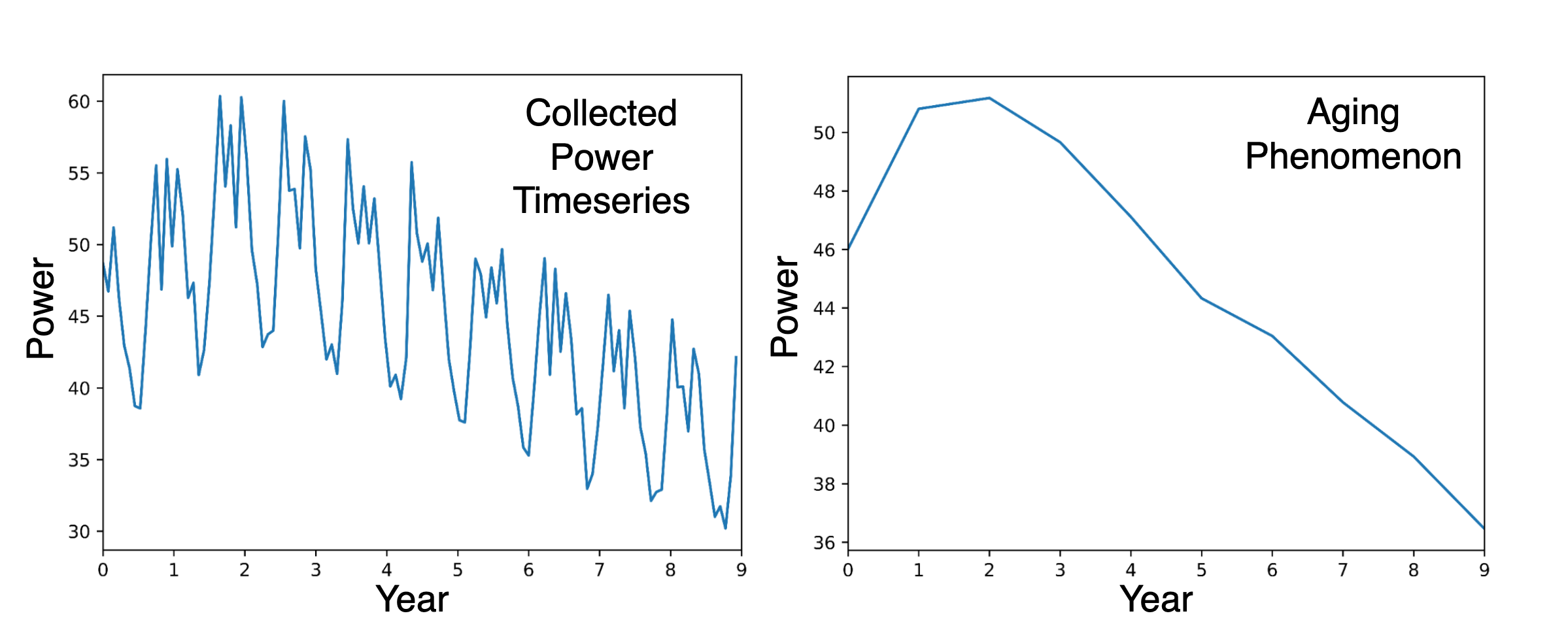}
  \vspace{-3ex}
  \caption{Example of a PV system with 10-years power timeseries exhibiting non-monotonic degradation pattern.}
  \label{fig:aging_visual}
  \vspace{-4ex}
\end{figure}

\vspace{.5ex}
Nevertheless, traditional PLR estimation typically models PLR as monotonic, linearly decreasing quantities, assuming linear degradation of the system components. In practice, PV degradation 
can be {\em neither linear nor monotonically decreasing}~\cite{hegedus2000analysis, jordanPVDegradationCurves2017}. This leads to erroneous estimation of the performance of PV systems over their lifetimes for methods using aforementioned assumptions \cite{jordanPVDegradationCurves2017}. 
An accurate understanding of the PLR of PV systems is critical to inform the LCOE calculation towards more accurate financial predictions, and allow for targeted assets maintenance ~\cite{delineReducingUncertaintyFielded2022b}.

\begin{example} 
\label{exa-nonlinear}
Fig. \ref{fig:aging_visual} shows example of a ten-years-long timeseries collected from a PV inverter. Observe that in first two years the performance actually goes up and then it degrades in a linear fashion till the end.
Hence, we cannot identify a global linear trend to characterize the performance loss rate for this inverter.

We observe that PLR in real PV systems do not always follow a monotonic decreasing or increasing trend. 
For example, the new installation 
of PV components may cause a break-in period when the performance of PV systems actually improves \cite{lindig2022best,jordanPVDegradationCurves2017}, then degrades as time goes by, and eventually shows up after the breaking point. 
Complex phenomena related to module operation can lead to cyclic patterns of power loss and recovery that are both technology and weather dependent.
Additionally, transient effects such as the build up of debris on module surfaces can intermittently lower power output \cite{meyersEstimationSoilingLosses2022}.
\end{example}

\vspace{.5ex}
While several data-driven approaches have been studied for PLR estimation, 
it remains to be a nontrivial task 
due to several computational challenges below.

\sstab
(1) {\em Lack of Large fleet-level Analysis}\label{C1}: Traditional PLR Power prediction methods such as 6K, PVUSA, XbX, and XbX+UTC~\cite{huld2011power,king1997field,curran2017determining,jordan2017robust} can only provide stand-alone PLR analysis for a single PV system, yet %cannot leverage spatial coherence 
does not scale to large-scale, fleet-level PLR analysis \cite{french2021assessment}.

\sstab
(2) {\em Domain-knowledge Heavy}\label{C2}: Conventional PLR estimation methods \cite{huld2011power,king1997field,curran2017determining,jordan2017robust} often rely on domain-knowledge based model tuning and ad-hoc decisions.
They rely on domain knowledge and manual effort, such as choosing appropriate thresholds, pre-defined filters and physical models. 
None of these approaches has addressed a streamlined automated PLR estimation pipeline that can ``cold-start'' without prior domain knowledge of PV timeseries data. This hinders large-scale PV PLR analysis.

\sstab
(3) {\em Diverse Degradation Patterns}\label{C3}: As remarked earlier, PLR degradation pattern can be non-linear and/or non-monotonic \cite{lindig2022best}, which cannot be extracted by degradation profiling approach that assumes the monotonic degradation pattern \cite{ulanova2015efficient}.
A single PLR value may fail to characterize non-linear PV degradation patterns. Degradation patterns of real-world PV systems may exhibit ``raise up then down'' rather than a monotonic change as time goes by.

%Observation 2
\vspace{.5ex} (Spatiotemporal) Graph Neural Networks 
%models and graph autoencoders (GAE) 
have been applied for non-linear timeseries forecasting,  
%and noisy data imputation tasks, 
with recent success in PV power predictive analysis, and PV data imputation~\cite{DBLP:journals/corr/abs-2107-13875,karimi2021spatiotemporal,10.1145/3588730}. 
These methods exploit spatio-temporal-topological features to better learn PV fleet-level 
representations, and have been 
verified to outperform state-of-the-art predictive methods in PV performance analysis. 

%%%%%%%%%
% we move these closer to justify 
% the architecture design section. 
%%%%%%%%%%
\eat{
Besides, from the view of representation learning, it is essential to learn disentangled seasonal-trend representations since separated parts can provide decomposition analysis and improve timeseries prediction~\cite{feng2021context}. Furthermore, instead of jointly encoding the multi-seasonal sub-series into a unified representation, ~\cite{woo2022cost} shows that multi-level disentangled seasonal levels each with a separate representation, are more robust to interventions such as the distribution drift (shifts in distribution of external factors).
}

\vspace{.5ex}
The above observations inspire us to consider new PLR estimation paradigms that take advantage of the following intuition. (1) In ideal PV systems without degradation, long-term power timeseries should be stationary, despite of the variations due to weather conditions across different seasons and years; and (2) power timeseries can be decomposed into two major components - a long-term ``degradation'' trend that represents the changes (``up'' and ``down'') in PV performance, and one or multiple fluctuation terms that capture the power timeseries seasonalities and noises.  We approach PV degradation estimation as an unsupervised regression problem and investigate GNN-basedd trend analysis to effectively captures long-term degradation patterns for large fleets of PV inverters.

\stitle{Contributions}.
We make the following  contributions.

\sstab 
(1) We propose \textbf{\stdyn}, a spatio-temporal graph autoencoder-based method that can effectively capture the degradation patterns in a fleet of PV systems.
\stdyn adopts paralleled graph autoencoders (one pattern per GAE) to decompose the input into separated trend and a series of different levels of fluctuation patterns and derives degradation from the trend (aging) GAE channel. 
\eat{\stdyn utilizes paralleled graph autoencoders to decompose input timeseries into aging and multiple fluctuation terms and uses the extracted aging terms to estimate degradation patterns.}

\sstab
(2) We design a unsupervised learning framework that does not need any prior domain-knowledge. 
We design a novel learning objective to ensure clear disentanglement between aging and fluctuation terms, consisting of three components: (1) Minimizing the reconstruction error between input and sum of aging and multiple fluctuation terms, (2) Ensuring the smoothness of aging term by reducing the noise level, and (3) Imposing the flatness conditions on the fluctuation terms to ensure stationarity of them.

\eat{\sstab 
(3) Our experimental results verify that \stdyn can cope with diversified degradation scenarios and datasets.
\stdyn achieves better degradation estimation across different degradation patterns and types of physical systems.}

\sstab 
(3) \stdyn supports scalable spatiotemporal graph learning and inference for long-term trend analysis. 
\stdyn achieves this through a novel top-down three-level ``Think-like'' parallelism graph learning algorithm \parastdyn. \eat{which consists of model parallelism (parallelizing each branch of GAE), data parallelism (parallelizing Snapshots), and pipeline parallelism (parallelizing ``Think-like layers'').}

To our knowledge, \stdyn is the first work that uses spatio-temporal graph autoencoders to perform fleet-level degradation analysis, addressing non-linear and non-monotonic 
trend analysis. 
We have deployed \stdyn on the CWRU CRADLE High Performance Cluster (HPC) for monitoring the degradation patterns of real-world PV systems using proprietary PV timeseries data from our PV manufacturer collaborators.
This enables us to provide them real-time feedback of the performance of their PV systems. \eat{add some discussion to highlight that \stdyn 
has been deployed at cradle hpc
and connected for real PV datasets. }

% \vspace{.2ex}
% To our knowledge, \stdyn is the first work that uses spatio-temporal graph autoencoders to perform fleet-level PLR analysis, addressing 
% non-linear and non-monotonic 
% trend analysis. 

\stitle{Related Work}. We summarize related work as follows. 

% \stitle{Related Work}. We summarize related work as follows. 
% \eetitle{Spatiotemporal Graph Neural Networks}.  
\eetitle{Spatiotemporal Graph Neural Networks}. 
Graph Neural Networks \cite{wu2020comprehensive} has been extensively investigated for graph representation learning. 
Spatio-Temporal graph neural networks (ST-GNNs) are an extension of GNNs that are designed to model the dynamic node inputs while assuming inter dependency between connected nodes \cite{wu2020comprehensive}. 
ST-GNNs capture both spatial and temporal dependencies of graph nodes with \eg recurrent graph convolutions  \cite{bai2020adaptive,seo2018structured} or attention layers \cite{Yu2018STGCN,zheng2020gman}. 
The former captures spatiotemporal coherence by filtering inputs and hidden states passed to a recurrent unit using graph convolutions, while the latter learns latent dynamic temporal or spatial dependency through attention mechanisms. 
%Such variants
%often better capture the topological information within PV networks than Convolutional Neural Networks (CNNs).

While prior study specifies GNN-based models for short-term time-series prediction (usually with pre-defined historical and predictive windows), 
not much has been investigated to capitalize ST-GNNs for providing a long-term trend analysis. Existing short-term predictive methods cannot be easily adapted for unsupervised, long-term PLR analysis, especially given non-linearity and non-monotonicity nature of the long-term trends. 

% \eetitle{Degradation Analysis in Physical Systems}. 
\eetitle{Degradation Analysis in Physical Systems}. In general, degradation models can be classified into experienced-based approaches, model-based approaches, knowledge-based approaches, and data-driven approaches \cite{gorjian2010review,shahraki2017review}. Neural Network-based data-driven approaches has become more popular due to their capacity of capturing complex phenomenon without prior knowledge and producing better approximation than traditional regression methods \cite{gorjian2010review}. 
Previous work \cite{ulanova2015efficient} proposes degradation profiling method for a single timeseries from the complex physical system and solve it as an optimization problem by quadratic programming. 
Albeit it achieves a clear separation between aging and fluctuation parts on a single timeseries, it fails to capture non-monotonic aging pattern due to the monotonic constraint imposed on aging part. In contrast, \stdyn leverages rich spatial correlation from a fleet of physical systems to extract degradation patterns for multiple timeseries, from multiple system components at once. 
Moreover, architecture design that supports 
effective scaling out with provable guarantees over large fleet level analysis is not addressed by 
prior work.

\vspace{-2ex}
\section{Preliminary}\label{sec:problem_statment}

\begin{table}[tb!]
\footnotesize
\caption{Summary of Notation.}
\vspace{-2ex}
% \begin{small}
\begin{center}
\begin{tabular}
{|c|c|}
\hline \textbf{Notation} & \textbf{Description} \\
\hline $G = (V,E,X_t)$ & a series of spatio-temporal graphs \\
\hline $G_t$ & snapshot of $G$ at time t \\
\hline $V$ & set of nodes in $G$ \\
\hline $E$ & set of edges in $G$ \\
\hline $A$ & adjacency matrix of $G$ \\
\hline $X_t$ & node attributes (timeseries)\\
\hline $k$ & the number of fluctuation terms \\
\hline $h_a$ & aging term \\
\hline $h_{f_i}$ & $i$-th fluctuation term \\
\hline $M$ & \stdyn Model \\
\hline $P$ & set of processors/workers \\
% \hline $\alpha$ & dissimilarity threshold in fine-tuning\\
% \hline $r$ & num. hops used to compute bisimulation\\
% \hline $EC$ & set of equivalent classes (maximum bisimulation)\\
\hline 
\end{tabular}
\end{center}
% \end{small}
% \caption{Summary of notation in \stdyn.}
\label{tab-notations}
\vspace{-4ex}
\end{table}

\stitle{PV Network Model}. We model a large-fleet PV timeseries dataset $\G$ = $\{G_1, \ldots G_T\}$ as a sequence of undirected graphs (``snapshots''), where 
for each snapshot $G_t = (V, E, X_{t})$ $(t\in[1,T])$, (1) each node in $V$ denotes a PV system or inverter; (2) $E$ refers to the edge set that
%thresholded gaussian kernels~\cite{shuman2013emerging} from Eq. \ref{second_Equation} to 
links (spatially correlated) PV systems; and (3) $X_{t} \in R^{N\times d}$ denotes the attribute/feature tensor at timestamp $t$, where 
(a) $N$ %is a %fixed constant number 
%that 
refers to the number of nodes in $G_t$ at time $t$; and (b) each node carries a $d$-ary tuple that records $d$ (PV) measurements at timestamp $t$. 
\eat{We denote the timeseries data 
(PV measurement) at node $v_i$ 
and timestamp t as $x_{i_t}$.}
%\warn{I have rephrased this part; and seems $T$ is not used or will be overloaded in parallel cost analysis, I removed it. please check.}
%Note 
%the total number of 
%time sequences $T$ is 
%|%x_1| = |x_2| = \ldots = |x_N|$.
%$d$ is 1 since each node carries one %power timeseries in our study.

\vspace{-1ex}
\begin{equation}
\label{second_Equation}
A_{i,j} = \begin{cases} 1, i\neq j \text{ and } \exp{(-\frac{dist_{ij}^2}{\sigma^2})} \geq \epsilon \\ 
0 \text{ otherwise } \end{cases} 
\end{equation}

% \vspace{1ex}
%The above equation determines 
%if two PV inverters should be linked. 
The above equation derives the adjacency matrix $A$ of a PV network. 
By default, \stdyn adopts $dist_{ij}$ as Euclidean distance between the locations\footnote{\stdyn supports other user-defined distance measures such as spatial-temporal correlation or 
feature differences as needed.} of PV systems $i$ and $j$. 
%\warn{this is not clear; the Euclidean distance of (i,j) -- defined with what? location? initial features? and how a gussian kernel is applied? clarify. }
$\sigma$ is standard deviation of all pairwise distances. 
$\epsilon$ is a threshold 
to control the network sparsity: the larger, the sparser the network is. 
\stitle{PLR: A Characterization}. 
% domain form, roger report inconsistent results
% Given a PV network $G$ with observed PV timeseries data $X = \{x_1, x_2,\ldots, x_N\}$ as input, our goal is to derive PLR estimation for each node (inverter) in a graph (fleet) with high accuracy . 
% We characterize the task as a unsupervised machine learning problem. 
% Given a set of observed power output timeseries, our goal is to learn multiple paralleled graph autoencoders that can decompose the input into aging $h_a$ and fluctuation terms $h_f$ separately by minimizing (a) the reconstruction loss between sum of all aging and fluctuation terms and original input $X$, (b) flatness regularization of $h_f$, and (c) smoothness constraint of $h_a$.
%
%traditional formulas try to approximately define PLR, Give an example, approximate PLR from domain-knowledge. Cite roger's report --> different methods give inconsistent results (here attack traditional methods again) -- even experts don't agree with each other on good PLR definition
%
%1: general characterization of PLR from domain
%
%2: model PLR be aging (trend) part regress analysis  (intuition (two parts like KDD), two parts like KDD)) on
%
%move to problem definition
PLR degradation pattern is %formally
defined as the ``performance loss''
measured by power output
%%%%%%%%%%%
% should this be performance or 
% performance loss?
%%%%%%%%%%%
of PV systems under the standard test conditions (STC) over time \cite{kaaya2020photovoltaic}.
STC~\cite{makrides2014performance} means fixed test conditions with $25\degree C$ cell temperature, $1000 W/m^2$ plane of module irradiance and AM $1.5G$ spectrum. 
PV scientists conduct accelerated indoor degradation testing and measure the power output under STC to estimate PLR.
Despite the simple computation, indoor testing may not reflect how the PV systems actually degrade in real-world due to the varying environmental factors to which real systems are exposed to.
This calls for an automated ML-based approach to estimate PLR of real systems using power output streams.
\eat{stadard test conditions 
is undefined here. Explain 
what it is and we need to  
clarify that if this is not computable without 
seeing the actual output; so to clarify ``why we need ML approach if there are already computable models for PLR?'' 
to make the motivation stronger.}
\eat{There are two approximating models to 
quantify PLR: relative PLR \cite{lindig2018review}, aggregated by comparing power data 
output to its counterpart from initial state of the PV system (with a unit  \%/annum ``\warn{add some explanation here: percentage annually? don't make people guess - and this unit occurred in Sec 1 end of first paragraph - so you can move this early in Sec 1 and remove it here}'', or simply \%/a); and absolute PLR \cite{ingenhoven2017comparison}, which is the coefficient of the slope 
of the power output as time goes by. 
Nevertheless, both assume a linear degradation rate, which cannot characterize diversified degradation patterns in real-world PV systems \cite{jordanPVDegradationCurves2017}. }
% It is a dynamic process that cannot be accurately characterized by a single PLR value.
We advocate a modeling of the  
%\warn{PLR degradation pattern}
degradation pattern to be the {\em aging (trend)} of the timeseries power output  
over time,  and derive the estimated degradation pattern after filtering out the fluctuation terms that capture seasonalities and noises.
%%%%%%%%%%%%%%%%%%%%%%%%%%
%%% the sentence below comes from no where and is logically broken; so 
% I moved it to closer to Eq 9. 
%%%%%%%%%%%%%%%%%%%%%%%%%%%
\eat{We can also implement regression analysis on the extracted aging to derive the global PLR value defined Eq. \ref{eighth_Equation}.
}
% (similar to aforementioned relative PLR).
%Next, we will introduce our formal problem definition.

\stitle{Problem Statement}. We 
model PLR estimation as a regression 
problem. 
% Consider a PV network with multiple power timeseries 
% data, the inputs of \stdyn include graph $G = \{G_1, \ldots, G_t,\ldots, G_T\}$ and they share the same adjacency matrix $A$.
% $N$ is the total number of nodes (inverters), and each $x_i$ represents the power timeseries of node $i$.
% $A$ is the adjacency matrix of PV network.
% We aim to decompose $X$ into aging term $h_{a}$ and $k$ different fluctuation terms $h_{f_1}, h_{f_1}, \ldots, h_{f_k}$ using $k+1$ GAEs. 
%input: raw PV data
%output: GNN-based model miniminze regression error for trend part
Given the raw PV power timeseries data from a PV fleet,
We train a graph neural networks-based model to minimize the error between the real degradation pattern and estimated degradation pattern, by solving an unsupervised regression problem.

More formally, consider a sequence of historically observed PV network graphs $G = \{G_1, \ldots, %G_t,\ldots, 
G_T\}$, 
where all graphs consist of the same set of nodes $V$ and share the same adjacency matrix $A$, with varying PV measurements at each node over time, 
we aim to obtain a St-GNN based model such that it fits the ``hidden'' PLR pattern by minimizing a loss function that 
quantifies two types of errors: (a) reconstruction error of the aging and fluctuation terms and (b) additional smoothness and flatness regularization.
% \begin{itemize}
%     \item reconstruction error of the 
%     aging and fluctuation terms;
%     \item additional smoothness and flatness regularization. 
% \end{itemize}

%reconstruction error to ensure high-quality of the decomposed signals, including both aging term and flctuation terms; 
%    \item difference between 
%\end{itemize} (2) smoothness regularization imposed on aging term by reducing the noise level, and (3)  regularization applied on the fluctuation terms to ensure stationarity.

\section{\stdyn Framework}\label{sec:methodology}

We next introduce \stdyn architecture and its components. 
% Our proposed framework can be used to monitor the performance of PV systems.

\stitle{Model Architecture}.
We start with the architecture of \stdyn. The model \stdyn consists of $k+1$ parallelized array of graph autoencoders (GAEs), which {\em decompose} input PV signals into 
one {\em aging term}, and $k$ {\em fluctuation terms}. 
(1) The first module (denoted as
$GAE_1$) extracts %, encodes and 
and learn an {\em aging} representation $h_a$, and
(2) each module $GAE_i$ 
($i\in [2, k+1]$) extracts a distinct fluctuation term.

\eetitle{Justification of design}. 
It is essential for \stdyn to learn  ``disentangled'' seasonal-trend representations. (1) Separatedly   
learned sub-patterns can drastically improve timeseries prediction~\cite{feng2021context}.
(2) Furthermore, instead of leaning a joint representation for multi-seasonal sub-series, \stdyn~{\em disentangles}~ seasonal informational using multiple GAEs: one for a separate fluctuation/seasonal representation. Such design has been shown to be more robust to interventions such as the distribution drift (shifts in distribution of external factors)~\cite{woo2022cost}. 
(3) Better still, 
the arrayed design 
enable a hybrid parallization 
scheme to scale the 
training and inference 
of \stdyn model, as verified 
in Section~\ref{sec:evaluation}. 

%Besides, from the view of representation learning, it is essential to learn disentangled seasonal-trend representations since separated parts can provide decomposition analysis and improve timeseries prediction~\cite{feng2021context}. Furthermore, instead of jointly encoding the multi-seasonal sub-series into a unified representation, ~\cite{woo2022cost} shows that multi-level disentangled seasonal levels each with a separate representation, are more robust to interventions such as the distribution drift (shifts in distribution of external factors).
\begin{figure*}[tb!]
\vspace{-2ex}
  \centering
\includegraphics[width=0.9\linewidth]{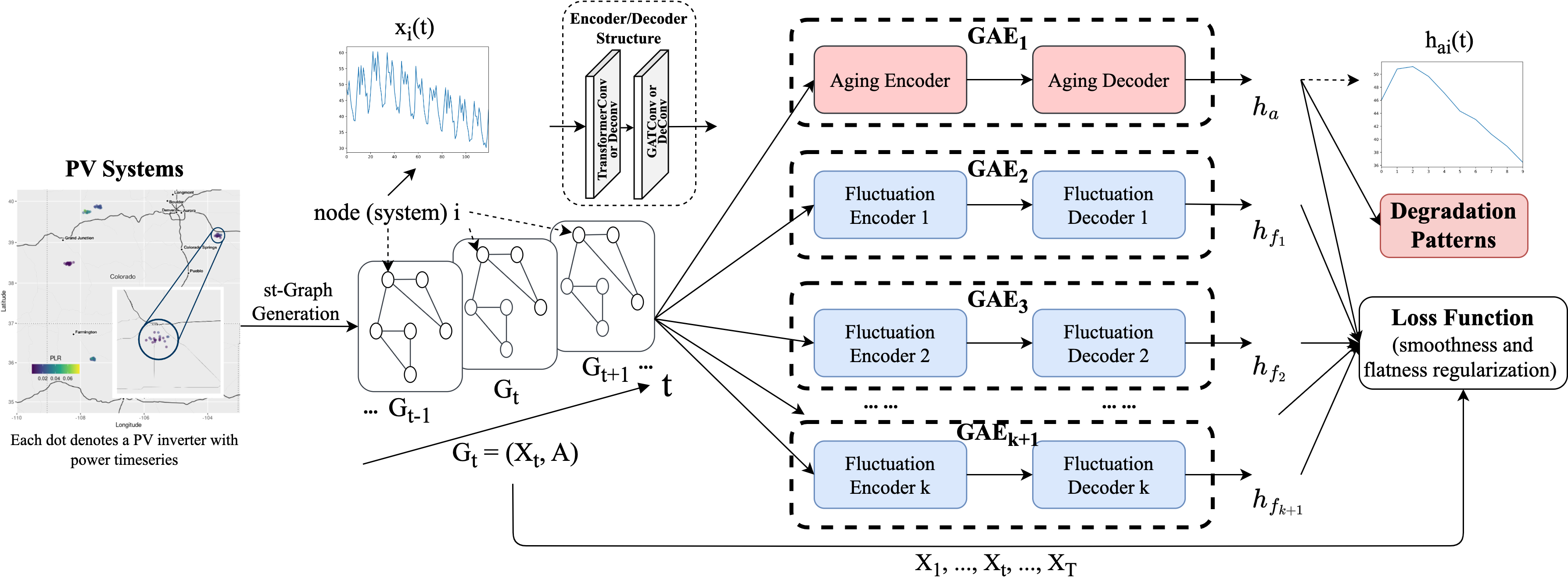}
  \vspace{-2ex}
  \caption{Overview of \stdyn Framework (six nodes shown in $G_{t}$ for illustration).}
  \label{fig_new:Framework}
\end{figure*}

We next specify our design of major 
components in \stdyn. 

\stitle{Parallel GAE Array}. As illustrated in Fig. \ref{fig_new:Framework},
Each GAE module consists of an encoder and decoder. 
 Each encoder or decoder consists of a graph transform operator (TransformerConv or TransformerDeConv) followed by a graph attention operator (GATConv or GATDeConv) to form two convolutional or deconvolutional layers. 
Both operators leverage attention mechanism. In graph transformer operator, the attention mechanism is applied independently to each node and involves self-attention across nodes.
On the other hand, graph attention operator computes attention coefficients using a shared self-attention mechanism across all nodes.
This design has its convention from 
established spatio-temporal GNNs \eg~\cite{Fey/Lenssen/2019}. 
We adapt such design for 
basic module in~\stdyn for 
aging and fluctuate 
representations for long-term 
trend analysis. 
 
\eetitle{Graph Transformer Operators}. Graph transformer operators, based on transformer architecture, perform attentive information propagation between nodes by leveraging multi-head attention into graph learning \cite{shi2020masked}.
Multi-headed attention matrices are 
adopted to replace the original normalized adjacency matrix as transition matrix for message passing.
Therefore, nodes can apply self-attention across nodes and update node representations to effectively aggregate features information from their neighbors.
We derive output of timeseries at node $i$ $x'_i$ as follows:  
% \begin{equation}
% \label{Eq2}
% \begin{split}
%     x'_i = W_1 x_i+\sum_{j \in N(i)}{\alpha_{i, j} W_2 x_j}; \\
%     \alpha_{i, j} = softmax(\frac{(W_3 x_i)^T (W_4 x_j)}{\sqrt{d}})
% \end{split}
% \end{equation}
\vspace{-1ex}
\begin{equation}
\label{Eq2}
\begin{split}
    x'_i &= W_1 x_i + \sum_{j \in N(i)}{\alpha_{i, j} W_2 x_j}; \\
    \alpha_{i, j} &= \text{softmax}\left(\frac{(W_3 x_i)^T (W_4 x_j)}{\sqrt{D}}\right)
\end{split}
\end{equation}

Here $W_1, W_2, W_3, W_4$ are trainable parameters, and $D$ is the hidden size of each attention head \eat{overloaded with d in Section 1? where d means feature length}.

\eetitle{Spatio-Temporal Graph Attention Operators}. 
We adapt the attention mechanism from graph attention networks (GATs) \cite{velickovic2017graph}. 
The operator uses masked self-attention layers and enable different weights to different nodes in a neighborhood by stacking layers, where nodes are able to attend over their neighborhoods’ features. This 
%mitigates the shortcomings in graph convolutions 
makes \stdyn more sensitive to 
useful information from 
spatial and temporal ``neighbors'' 
in the PV networks. 
% \warn{TBF: Formula for Graph Attention Operators}.
% \begin{equation}
% \label{Eq3}
% \begin{split}
%     x'_i = \sigma(\sum_{j \in N(i)}{\alpha_{i,j} W h_j}); \\
%     \alpha_{i,j} = \frac{\exp{\text{LeakyReLU}(a^T[W h_i||W h_j])}}{\sum_{k \in N(i)}{\exp{\text{LeakyReLU}(a^T[W h_i||W h_k])}}}
% \end{split}
% \end{equation}
We derive the updated representation of timeseries at node $i$ $x'_i$ as follows: 
\vspace{-1ex}
\begin{equation}
\label{Eq3}
\begin{split}
    x'_i &= \sigma\left(\sum_{j \in N(i)}{\alpha_{i,j} W h_j}\right); \\
    \alpha_{i,j} &= \frac{\exp(\text{LeakyReLU}(a^T[W h_i||W h_j]))}{\sum_{j \in N(i)}{\exp(\text{LeakyReLU}(a^T[W h_i||W h_j]))}}
\end{split}
\end{equation}

Here $a$ is the shared learnable attention parameter vector, $W$ is learnable weight matrix, $h_i$, $h_j$ are input timeseries for nodes i and j, respectively. $\sigma$ is an activation function.

\eat{
due to the high learning capacity of GAEs, it may be hard to achieve a clear disentanglement among the output of each GAE. 
Next, we will introduce how we define our loss function such that it can achieve a separation between aging and fluctuation.
}

%\vspace{-1ex}
\stitle{Learning Objective}. 
\stdyn introduces rich expressiveness 
with multiple parallelized GAE array. %On the other hand, 
Meanwhile, 
a clear disentanglement among 
GAE outputs should be 
achieved to distinguish 
aging and fluctuation 
representations. 
We next introduce  
a novel design of 
the loss function such that 
it can achieve a separation between 
aging and fluctuation 
for long-term trend analysis 
for PLR estimation. 
The loss function %of \stdyn 
consists of reconstruction error, flatness regularization, and smoothness regularization.

\eetitle{Reconstruction Error}. We aim to decompose $X$ into aging term $h_{a}$ and $k$ different fluctuation terms $h_{f_1}, h_{f_1}, \ldots, h_{f_k}$ using $k+1$ GAEs. To ensure the quality of decomposition, we want to minimize the reconstruction error as follows:
\vspace{-1ex}
%\begin{small}
\begin{equation}
\label{Eq4}
% X_{rec} = h_{a} + \sum_{q=1}^{k} h_{f_q}
% \end{equation}
RE = ||X-h_{a}-\sum_{q=1}^{k} h_{f_q}||^2 
\end{equation}
\vspace{-1ex}
%\end{small}

\eetitle{Flatness Regularization}. To ensure stationarity of fluctuation, we propose two constraints on $h_f$: (1) mean constraint: we segment every timeseries to ensure difference of mean between each segment within the series being as small as possible and (2) slope constraint: sum of absolute value of global slope of extracted $h_f$ for each node should be as close to zero as possible.

%\eetitle{Mean Constraint}. 
For mean constraint (MC), we partition $h_f$ of each node into $p$ segments each with length $w$. The length of w determines the temporal resolution of extract fluctuation term. We then minimize the sum of difference between mean values of each pair of the segments to ensure flatness of the fluctuation over time defined as:
\vspace{-1ex}
\begin{equation}
\label{Eq5}
MC = \sum_{q=1}^{k} \sum_{l=1}^{N} \sum_{i,j=1}^{p} (m_{q_{l_i}}-m_{q_{l_j}})^2 W_{ij}
\end{equation}
Here (1) $m_{q_{l_i}}$ denotes the mean of the $i$-th segment of the $l$-th node's $q$-th fluctuation term, (2) $W$ is a weight matrix, where each entry $W_{ij}$ denotes a learnable weight to minimize the mean difference between segmented pair $m_{l_i}$ and $m_{l_j}$. 
To ensure the long-term mean being minimized, we apply linear growth of the weight based on the distance between $m_{l_i}$ and $m_{l_j}$. This is based on the following intuition: the farther the two segments are from each other, the more weights are given to minimize their mean difference. 

\eetitle{Slope Constraint}. The slope constraint (SC) ensures the global flatness of each fluctuation level, and is defined as:
\vspace{-1ex}
\begin{equation}
\label{Eq6}
SC = \sum_{q=1}^{k} \sum_{l=1}^{N} \big | slope(h_{f_{q_l}})\big |
\end{equation}
Here the slope is calculated from the least square fit.

\eetitle{Smoothness Regularization}. We also want to reduce noises in $h_a$ by minimizing sum of the standard deviation of the first-order differences of aging term of all the nodes. We thus introduce a  smoothness regularization (SR) term as:
\vspace{-1ex}
\begin{equation}
\label{Eq7}
SR = \sum_{l=1}^{N} SD(h_{a_l}[t + 1] - h_{a_l}[t])
\end{equation}
where $t \in [1,T-1]$ with $T$ the total number of %$h_{a_{l}}$
timestamps, and SD denotes standard deviation.
% Note that the length of w determine the temporal resolution of fluctuation levels. Due to the nature of PV timeseries and resolution of collected timeseries data, we choose

\eetitle{Loss Function}. Putting these together, we formulate the loss function of \stdyn as:
% \vspace{-2ex}
% \begin{small}
% \begin{equation}
% \label{loss}
% \begin{split}
% % \begin{aligned}
% % \begin{multline*}
% \min ||X-h_{a}-\sum_{q=1}^{k} h_{f_q}||^2 + 
% \lambda_1 \sum_{q=1}^{k} \sum_{l=1}^{N} \sum_{i,j=1}^{p} (m_{q_{l_i}}-m_{q_{l_j}})^2 W_{ij} \\
% + \lambda_2 \sum_{q=1}^{k} \sum_{l=1}^{N} \big | slope(h_{f_{q_l}})\big | +
% \lambda_3 \sum_{l=1}^{N} SD(h_{a_l}[t + 1] - h_{a_l}[t])
% % \end{multline*}
% % \end{aligned}
% \end{split}
% \end{equation}
% \end{small}
\vspace{-0.5ex}
\begin{equation}
\label{loss}
% \begin{aligned}
% \begin{multline*}
\mathcal{L}(X,h_a,h_f) = RE + \lambda_1 MC + \lambda_2 SC + \lambda_3 SR
% \end{multline*}
% \end{aligned}
\end{equation}
Here RE, MC, SC, SR are defined in the Eq.~\ref{Eq3}-\ref{Eq6}. $\lambda_1$, $\lambda_2$, and $\lambda_3$ control the trade-off between reconstruction error and the quality of disentanglement between aging and fluctuations. The loss function aims to minimize the reconstruction loss while ensuring the flatness of the fluctuation terms and smoothness of the aging term.

%We can also implement regression analysis on the extracted aging to derive the global PLR value defined Eq. \ref{eighth_Equation}.

\stitle{PLR Calculation}. 
The output of \stdyn can be used to derive both estimated degradation pattern (EDP) and global PLR. 
$h_{a}$ directly provide the EDP for all systems in the PV fleet.
\eat{It can also be used to derive the global PLR estimation for each PV system.}
We derive global PLR of system $i$ by calculating the percent of performance loss or gain between any element $v$ in $h_{a_i}$ and its corresponding one year apart element $v'$ through:
\begin{equation}
\label{eighth_Equation}
PLR_{global} = \frac{1}{m} \sum_{i=1}^{m} \frac{v'-v}{v} \times 100\%
\end{equation}
Where $m$ is the total number of one year apart pairs in $h_{a_i}$.

% \subsection{Encoder and Decoder Structure}\label{sec:encoderr_decoder_structure}
% We next detail our design of encoder and decoder in GAE.
% As illustrated in Fig. \ref{fig_new:Framework}, every encoder or decoder in GAE consists of a graph transform operator (TransformerConv or TransformerDeConv) followed by a graph attention operator (GATConv or GATDeConv) \cite{Fey/Lenssen/2019} to form two convolutional or deconvolutional layers.

% Graph transform operator performs attentive information propagation between nodes by leveraging multi-head attention into graph learning \cite{shi2020masked}.
% Multi-head attention matrix replaces the original normalized adjacency matrix as transition matrix for message passing.
% Therefore, each node can effectively aggregate both features information from its neighbors.

% Graph attention operator is adopted from graph attention networks (GATs), a novel attention-based neural network architectures that operate on graph-structured data \cite{velickovic2017graph}. 
% It leverages masked self-attention layers to
% address the shortcomings and key challenges of methods based on graph convolutions, or their approximations. 
% It enables specifying different weights to different nodes in a neighborhood by stacking layers where nodes are able to attend over their neighborhoods’ features.
% In this way, the spatial correlation between node (inverters) in PV network can be better modeled instead of merely assigning equal weight to the neighbors.

\stitle{Time-cost Analysis}.
%optional time-cost analysis (paralled friendly)
We present an analysis on the 
inference cost of \stdyn. 
It takes in total $O(Lmf + LPnf^2)$ \cite{gao2022efficient} 
to derive decomposed results for a timestamp. Here 
$n$, $m$, $L$, $P$ and $f$ denote number of nodes, number of edges, number of layers of GAE modules, propagation order of message passing (which is bounded by the diameter of  PV network), and dimensions of the node features, respectively.
As $L$ and $P$ are often small 
for practical GAE designs, and $f$ are small constants in real PV data, the overall inference cost (including estimating EDP or global PLR) is $O((Lmf + LPnf^2)T + nT)$, which is 
bounded in $O((m+n)T)$.

% \eetitle{Parallel-friendly}. 
% \subsection{Parallel-friendly Model Deployment}\label{sec:model_deploy}

% \stitle{Parallel-friendly Model Deployment}.

\begin{figure}
\centering
\begin{algorithm}[H]
\caption{: \parastdyn}
% \label{alg:forward}
\begin{algorithmic}[1]
\algtext*{EndFor}
\algtext*{EndIf}
\algtext*{EndWhile}
\algtext*{EndFunction}
\algtext*{EndProcedure}

\State \textbf{Input:} 
A batch of snapshots $\G = \{G_1, \ldots, G_T\}$, (optional) \stdyn Model $M$, % = \{GAE_1, GAE_2, \ldots, GAE_{k+1}\}$, 
number of fluctuation terms $k$, a set of window sizes $W$, a set of processors $P$, a coordinator processor $P_0$, number of epochs $e$;
\State \textbf{Output:} 
Incrementally trained $M'$ upon batch $\G$.
\vspace{1ex}
% \State I
\For{$m = 1$ \textbf{to} $e$}
    \For{$i$ in $1$ \textbf{to} $k+1$} \textbf{in parallel} 
        \Comment{Model Parallelism}
        \If{$m = 1$}
            \State $P_0$ creates $G_i$, a copy of $G$;
        \EndIf
        \For{$j$ in $1$ \textbf{to} $\lceil \frac{T}{W_i} \rceil$} \textbf{in parallel} 
            \Comment{Data Parallelism}
            \If{$m = 1$}
                \State $P_0$ ships $G_{i, j} = G_{j+(i-1) \times W_i:j+i \times W_i}$ to $P_{i, j}$;
                \State $P_{i, j}$ creates $GAE_{i, j}$; 
            \EndIf
            \State $GAE_{i, j}.forward(G_{i, j})$; \Comment{Pipeline Parallelism}
            \State Backpropagate and update local gradient of $GAE_{i, j}$;
        % \EndFor
    \EndFor 
    \If{$m = epochs$}
            \State $GAE_i = aggregate(GAE_{i, j}), \forall j \in [1, \lceil \frac{T}{W_i} \rceil]$;
    \EndIf
\EndFor 
\EndFor
\State $P_0$ derives $M'$ by assembling all $GAE_i$, $\forall i \in [1, k+1];$
\State \Return $M'$ from $P_0$; 
\end{algorithmic}
\end{algorithm}
\vspace{-2ex}
\caption{\parastdyn: Three-level Parallel Training.}
\label{fig:para}
\vspace{-4ex}
\end{figure}

\section{Parallelization and Deployment}

\begin{figure}[tb!]
    \centering
    \includegraphics[width=\linewidth]{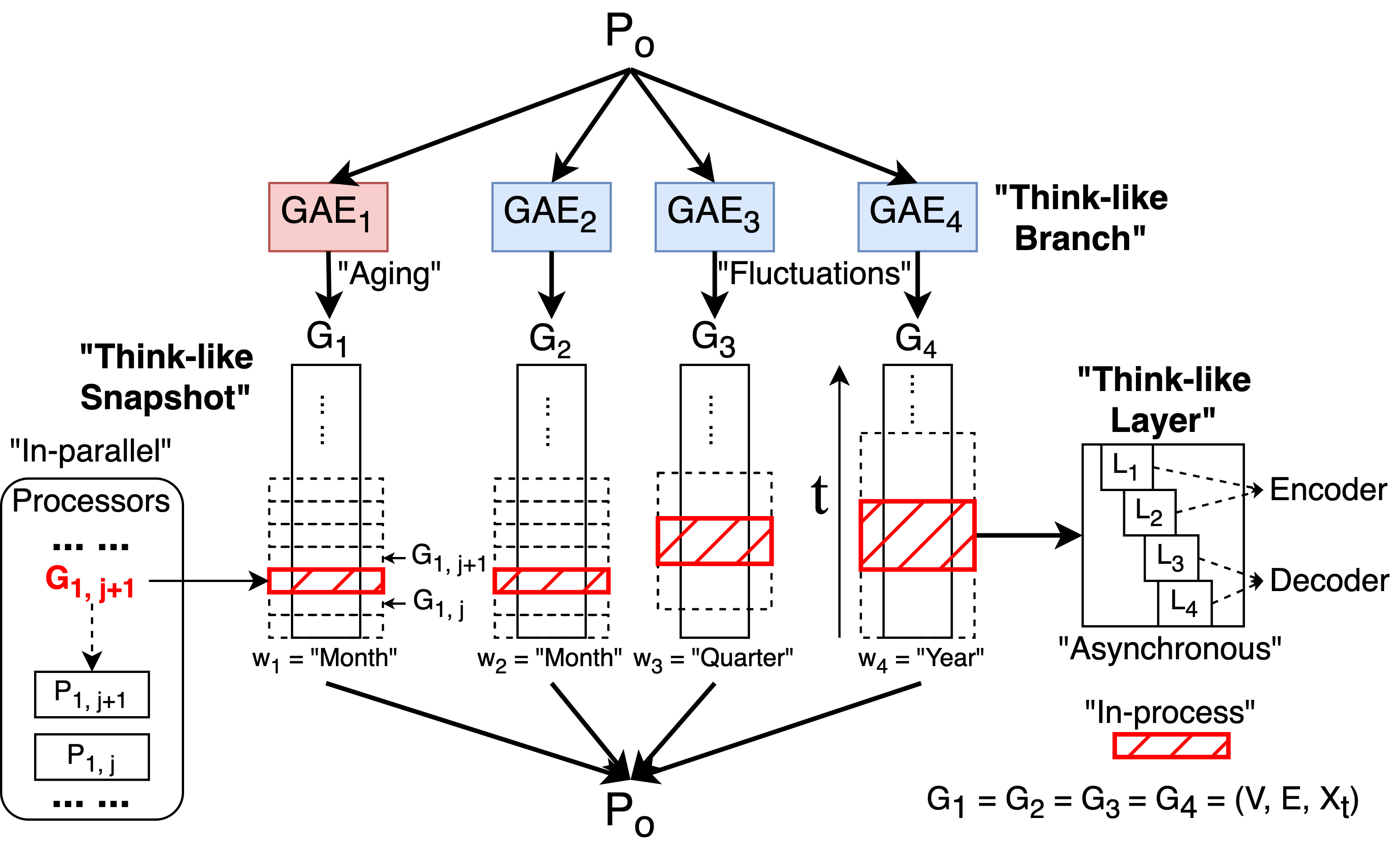}
    \caption{Proposed \parastdyn ($k = 3$ for illustration: one aging channel and three fluctuation channels).} 
    \label{fig:para_illustration}
    \vspace{-4ex}
\end{figure}

\eat{The design of \stdyn is ``parallel-friendly''~\cite{ben2019demystifying}.
For scalable training, we ship $k+1$ copies of the input PV data to each GAE module and perform a parallel computing at $k+1$ processors to generate embeddings and derive decomposed results simultaneously at a coordinator following \eg parameter server~\cite{li2013parameter}. 
Due to the limited space, we defer the study of large-scale \stdyn training and deployment in future work.}
The design of \stdyn is ``parallel-friendly''~\cite{ben2019demystifying}.
We next introduce a parallel algorithm, denoted as \parastdyn, to scale the  learning and inference of \stdyn to large PV network degradation analysis. 
%, as shown in Fig.~\ref{fig:para}. 
%(algorithm) and ~\ref{fig:para_illustration} (illustration).
\parastdyn exploits 
parrallelism with three levels of 
computation (as illustrated in Fig.~\ref{fig:para_illustration}).

\eetitle{Level I: ``Think-like Branch'' Model Parallelism}. \stdyn contains parallelized GAE array with $GAE_1, \ldots, GAE_{k+1}$ branches, as shown in Fig.~\ref{fig_new:Framework}.
$GAE_1$ is the aging branch.
$GAE_2, \ldots, GAE_{k+1}$ are the fluctuation branches. 
This presents opportunities for parallelizing the 
training by distributing GAE branches among  processors. In each training epoch, \parastdyn computes the forward propagation in parallel without coordination. 
The output of each branch will be assembled together by an coordinator processor $P_0$ to calculate global loss in Eqn.~\ref{loss}.
Then each branch backpropagates independently and updates their local gradients in parallel.

\eetitle{Level II: ``Think-like Snapshot'' Data Parallelism}. Within 
each branch, \stdyn takes a sequence of ``snapshots'' $\{G_1, \ldots, G_T\}$ as the input. \parastdyn processes the temporal components of the input with a specified length $L$ in parallel. 
For each fluctuation branch, $L$ is set to be the window size $w$ for the corresponding fluctuation branch.
For example, in Fig.~\ref{fig:para_illustration}, there are three fluctuation branches, each with different window size corresponding to ``month'', ``quarter'', and ``year''.
For aging branch, $L = \min_{j \in [2, k+1]} w_j$.
Besides, \stdyn exploits the mini-batch data parallelism (MBDP) in DGL~\cite{zheng2020distdgl} to achieve even larger speed-up. 
It splits the information propagation of the network into parallelly computed message flow graphs induced by mutually disjoint node batches.

\eetitle{Level III:``Think-like Layer'' Pipeline Parallelism}. Each branch of \stdyn encompasses two layers for encoder, and another two layers for decoder. 
\eat{Since layers in decoder and encoder of each GAE are symmetric to each other, the forward/back propagation of encoder is equivalent to the back/forward propagation of decoder.} 
We adopt an asynchronous macro-pipeline parallelism schema~\cite{narayanan2019pipedream} to parallelize the computation of layers within each GAE, such that inter-layer synchronization is eliminated (without information loss, \ie the batched graphs from level II are independent of each other).
A worker processing its graph batch starts the graph operation on the new layer as soon as it completes the non-linearity in previous layer~\cite{besta2022parallel}.
% In future work, We plan to develop an asynchronous macro-pipeline parallelism schema to parallelize the computation of layers within each GAE.

To measure the parallelism of \parastdyn, 
we extend the notion of 
parallel scalability~\cite{kumar1991analysis,fan2014distributed} to 
graph learning. 

\begin{definition}
A training algorithm is {\em parallel scalable} if its parallel time cost is inverse proportional to the number of parallel processors, with additional cost that is independent of input size. 
%approaching the theoretical optimal speed-up. 
\end{definition}

% \begin{figure}
% \centering
% \begin{algorithm}[H]
% \caption{: \parastdyn}
% % \label{alg:forward}
% \begin{algorithmic}[1]
% \algtext*{EndFor}
% \algtext*{EndIf}
% \algtext*{EndWhile}
% \algtext*{EndFunction}
% \algtext*{EndProcedure}

% \State \textbf{Input:} 
% Spatio-temporal Graphs $G = (V, E, X_t)$, \stdyn Model $M = {GAE_1, GAE_2, \ldots, GAE_{k+1}}$, number of fluctuation terms $k$, a set of window sizes $W$, a set of processors $P$, a coordinator processor $P_0$;
% \State \textbf{Output:} 
% Trained $M'$ using $G$.
% \vspace{1ex}
% % \State I
% \For{$i = 1$ \textbf{to} $k+1$}
% \Comment{Model Parallelism}
%     \For{$j = 1$ \textbf{to} $\lceil \frac{T}{W_i} \rceil$}
%         \Comment{Data Parallelism}
%         \State $P_0$ ships $G_{i, j} = \{G_{j+(i-1) \times W_i}, \ldots, G_{j+i \times W_i}\}$ to $P_{i, j}$;
%         \State create $GAE_{i, j}$, a copy of $GAE_i$ in $P_{i, j}$; 
%         \State train $GAE_{i, j}$ using $G^{i, j}$; \Comment{Pipeline Parallelism}
%     \EndFor
%     \State $P_0$ assemble all $GAE_i$, $\forall i \in [1, k+1];$
% \EndFor 
% \State \Return $M'$ from $P_0$; 
% \end{algorithmic}
% \end{algorithm}
% \caption{\parastdyn: Three-level Parallelism Paradigm for \stdyn.}
% \label{fig:para}
% \end{figure}

\begin{theorem}
\label{th1} 
The training algorithm 
\parastdyn is scale-free with a total parallel cost in $O\left(\frac{T(G, M)}{\lvert P \rvert} + f(\theta)\right)$ time, where $T(G, M)$ is total cost without parallelism and $f(\theta)$ is independent of the size of $G_t$ and linear to the length of timeseries.
\end{theorem}

\eetitle{Proof Sketch}. Given the input G (a series of $G_t$) and model M and the total cost of \stdyn training without parallelism $T(G, M)$, since the maximum speed-up is bounded by the number of processors/workers utilized, which is $\lvert P \rvert$, the lower bound of the cost of \parastdyn is $\frac{T(G, M)}{|P|}$.
We denote the communication overhead between the coordinator and processors/workers as $f(\theta)$ which is in $O(k \lceil \frac{T}{W_{min}} \rceil e)$, where $w_{min} = \min\{w_1, w_2, \ldots, w_{k+1}\}$. 
Since $w_{min}$ and $k$ are hyper-parameters, $f(\theta)$ is independent of the the number of nodes $|V|$ and number of edges $|E|$.
$f(\theta)$ is linear to the length of timeseries $|T|$.
Therefore, the total cost of \parastdyn is in $O\left(\frac{T(G, M)}{\lvert P \rvert} + f(\theta)\right)$ such that $f(\theta)$ is independent of size of the $G_t$ and linear to the length of timeseries.

\begin{figure}[tb!]
    \centering
    \includegraphics[width=0.95\linewidth]{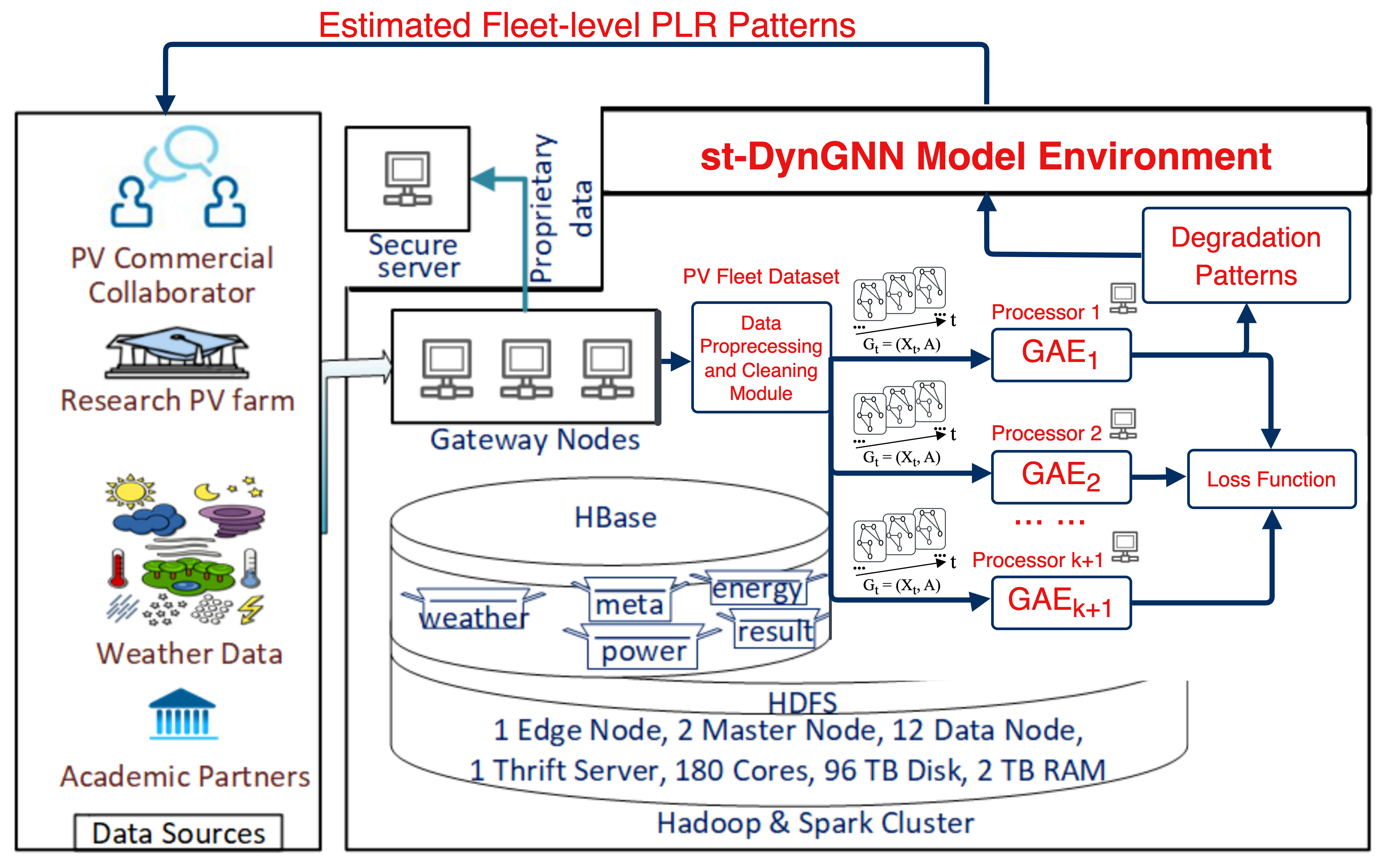}
    \vspace{-2ex}
    \caption{Illustration of \stdyn Workflow with Level I Parallelism Implemented for PV Systems from Our Industry Partners (deployed in CRADLE~\cite{nihar2021toward}). \eat{A cron job periodically executes a program to ingest timeseries data from different PV inverters into Hadoop cluster. $k+1$ GAE modules receive copies of data at same time and extract aging and $k$ fluctuation terms. PLR results derived from aging term are then shipped to collaborators.}} 
    \label{fig:Data_flow}
    \vspace{-4ex}
\end{figure}

\stitle{System Deployment}. \stdyn has been deployed to improve PV PLR estimation for commerical plants from SolarEdge (energy company).
Fig.~\ref{fig:Data_flow} illustrates how \stdyn (with Level I Model Parallelism illustrated) is  incorporated into the our high performance cluster (HPC) CRADLE~\cite{nihar2021toward} and communicates with industrial partners. 
Acquisition of high-quality data is challenging and crucial for PLR degradation analysis on real-world PV systems.  
Our developed spatio-temporal data imputation python package~\cite{PVplr2024} provides high-quality data for \stdyn training and inference. 
\stdyn leverages \parastdyn to accelerate training and inference using computational resources in HPC.
The estimated fleet-level PLR results are then forwarded to our PV commercial collaborators and research PV farms.
Our proposed framework is generally applicable to other domains that involves spatiotemporal long-term timeseries analysis,\eg trend analysis in traffic volumes at intersections, air quality changes in urban environments, monitoring the trend of the spread of infectious diseases, and etc. Domain knowledge from corresponding fields can guide the choice of hyperparameters such as the number of fluctuation terms $k$ and corresponding window sizes $w$ needed and trains \stdyn accordingly.

\section{Experiments}\label{sec:evaluation}

We experimentally verify the 
performance of \stdyn compared with baselines, in terms of accuracy and case analysis of estimated degradation patterns.

\subsection{Experimental Setup}\label{sec:config}
\stitle{Evaluation Metrics}. 
We evaluate the degradation estimation errors using Euclidean Distance (ED) and Mean Absolute Percent Error (MAPE).
Smaller values of ED and MAPE indicate better PLR estimations, as they represent a closer match between the EDP (estimated degradation pattern) and RDP (real degradation pattern).

\eetitle{ED}. 
Reporting a single degradation rate for the entire system does not capture the system's performance dynamics over time. 
We establish a novel way to quantify how well a model can estimate PLR degradation pattern, consisting of the following two steps: (1) Rescaling: Given RDP and EDP, we rescale every data point in RDP and EDP by dividing them by their respective first values and (2) Error Calculation: We calculate ED between scaled RDP and EDP.

\eetitle{MAPE}. 
We define MAPE as follows:
\begin{equation}
\begin{aligned}
% MAE = \frac{1}{N \times T} \sum_{L=1}^{N} \sum_{J=1}^{T} |EDP_{L_J}-RDP_{L_J}|; \\
MAPE = \sum_{L=1}^{N}  \sum_{J=1}^{T} \frac{|EDP_{L_J}-RDP_{L_J}|}{N \times T \times |RDP_{L_J}|} \times 100\%
\end{aligned}
\end{equation}
% h_{f_q}
where $EDP_{L_J}$ is the $J-th$ coefficient of EDP for node $L$.

\stitle{Datasets}.\label{dataset}
We verify our model on five datasets, including three PV datasets and two real-world datasets from Economy and Finance: (1) \textbf{PV\_Case1}: It refers to the PV datasets exhibiting global linear degradation patterns. (2) \textbf{PV\_Case2}: It refers to the PV datasets showing a piecewise linear degradation pattern with a breakpoint (a change in the degradation rate at the second year). 
(3) \textbf{PV\_Case3}: This refers to the PV dataset exhibiting a non-linear exponential degradation. 
Each PV dataset consists of 10-year power output timeseris from 100 inverters in 5 PV sites from Colorado, USA.
Every power output timeseries includes sample interval of 15 minutes, amounting to 350,592 data points for each PV inverter. 
Each PV site contains spatially correlated inverters such that inverters in close proximity experience similar degradation severity. 
Each inverter (PV system) is located in one of the five clusters, each with a $\pm 4\%$ noise in geo-spatial position, and with a $\pm .2\%$ noise.

To demonstrate the generality of 
\stdyn for trend analysis in other 
applications, we also adopt the 
following public datasets. 
(4)
\textbf{Finance}~\footnote{https://finance.yahoo.com/}: This refers to weekly stock prices of the 500 companies listed in S\&P 500 from 12/13/2018 to 12/13/2023.
(5) \textbf{Economy}~\footnote{https://data.worldbank.org/}: 
%This refers to 
an annual GDP (in current US dollars) of G20 countries in past 50 years sampled from 1973 to 2022. 
For finance (resp. economy) data, 
each node represents a 
stock (resp. country). For both datasets, 
%a different edge construction approach is employed. 
we used a thresholded pairwise absolute pearson correlation among node attributes (timeseries) to establish edges. 
%with $\epsilon$ as a threshold in the construction of $A$.
% \eetitle{Hydrology}~\footnote{https://nrfa.ceh.ac.uk/}. This refers to NRFA Peak Flow Dataset. It contains annual peak flow of rivers timeseries derived from 917 
% flow gauging stations across the UK.

We obtain the RDP of stock and GDP timeseries using empirical mode decomposition~\cite{huang1998empirical} which is widely applied in finacial and economic trend analysis~\cite{deng2022multi, dai2020forecasting, mao2020analysis, saadaoui2020multiscaled}. 
We split each dataset by nodes, 50\% for training, 25\% for validation, and 25\% for testing.

\stitle{Baselines}. 
% We compare \stdyn with four state-of-the-art PLR Power Performance Models.
We compare \stdyn with eight baselines.
% \warn{make the style consistent.}

\sstab 
(1) \underline{{\em 6K}} \cite{huld2011power}: 
a domain-specific model, which incorporates irradiance and module temperature as a fraction of standard irradiance and difference from standard temperature. 
% Additionally, this model requires a nameplate power input and will always predict the supposed nameplate power at Standard Test Conditions (STC).

\sstab
(2) \underline{{\em PVUSA}} \cite{king1997field}: 
an established physics-based PLR estimation model.
The assumption of the model is that the current of a solar panel is a function of the irradiance and the voltage is a function of the irradiance and the module temperature.
% , which is predicted by the ambient temperature and the wind speed. 

\sstab
(3) \underline{{\em XbX}} \cite{curran2017determining}:  a data-driven, multiple regression predictive model. The model enables change of point PV degradation pattern modeling. 
% The X in the name refers to a given time step over which the power prediction model is built over; a model built on a day of data would be Day-by-Day (DbD), while in Month-by-Month (MbM) modeling, data would be subset by months. The time step is chosen based on the condition of the data being modeled, and what modeling will be performed on the overall dataset. 

\sstab
(4) \underline{{\em XbX + UTC}} \cite{jordan2017robust}: The XbX + UTC is based on the XbX model by introducing a universal temperature correction (UTC) to produce a single temperature coefficient that can be used to convert to the desired representative temperature value.

\sstab
(5) \underline{{\em MSTL}} \cite{trull2022multiple}: 
Multiple Seasonal-Trend Decomposition using Loess (MSTL) decomposes a time series into a: trend component, multiple seasonal components, and a residual component. 
MSTL uses STL to iteratively extract seasonal components from a time series. 

\sstab
(6) \underline{{\em SSA}} \cite{apaydin2021artificial}: 
Singular Spectrum Analysis (SSA) allows extraction of alleged trend, seasonal and noise components from time series. 
The name comes from singular decomposition of a matrix into
its spectrum of eigenvalues. 
The time series can be reconstructed
by regrouping different important components.

\sstab
(7) \underline{{\em QP-Aging-Detect}} \cite{ulanova2015efficient}: 
The QP-Aging-Detect, is a mathematical model using Quadratic Programming (QP) to profile long-term degradation by decomposing timeseries into the aging and fluctuation terms. 
For fair comparison, we remove the monotonic (strictly increasing or decreasing) constraint imposed on aging component when implementing it in CVXPY \cite{diamond2016cvxpy}.

\sstab
(8) \underline{{\em STGAEs}}: 
Spatio-Temporal Graph Autoencoders (STGAEs) are variants of \stdyn, where the encoder and decoder layers in \stdyn are replaced by either Graph Convolutional (GCN) Layers~\cite{kipf2016semi} for STGAE1 or graph attention layers~\cite{velickovic2017graph} for STGAE2.

% \begin{figure}[tb!]
%     \centering
%     \includegraphics[width=\linewidth]{fig_new/Hyperparameter.png}
%     \vspace{-3ex}
%     \caption{The Impact of Epsilon (the larger epsilon the sparser the graph) and Number of Epochs on PLR Degradation Estimation Error in Case 3 PV Dataset.}
%     \label{fig:tuning}
%     \vspace{-3ex}
% \end{figure}

\stitle{Hyperparameter Tuning}. 
We use grid search based on the validation loss to identify the optimal set of hyper-parameters \cite{pontes2016design} for \stdyn.
We varied the number of fluctuation term $k$ in the set $\{1, 2, 3, 4\}$, network sparsity $\epsilon$ in the set $\{0, 0.25, 0.5, 0.75, 1\}$, number of epochs in the set $\{100, 250, 500, 750, 1000\}$, learning rate from $\{0.001, 0.01, 0.05, 0.1\}$, regularization terms $\lambda_1$, $\lambda_2$, and $\lambda_3$ all in the set $\{1, 5, 10, 25, 50, 75, 100, 200\}$.
Based on the validation loss, we choose $k$ equal to 1, $\lambda_1$, $\lambda_2$, and $\lambda_3$ to be 5, 100, and 10 separately, epsilon to be 0.5, epochs to be 500, the learning rate to be 0.05. 
% In Fig.~\ref{fig:tuning}, we use case three degradation pattern dataset as an example to illustrate how error (MAPE) change when we vary either epsilon or epochs in training.
% Based on above analysis, we select epsilon to be 0.5 and epochs to be 500.

\stitle{Model Training}. \stdyn is trained by Adam optimizer \cite{kipf2016semi} and implemented with PyTorch Geometric Temporal \cite{rozemberczki2021pytorch}, on 12 Intel(R) Xeon(R) Silver 4216 CPU @ 2.10GHz, 128 GB Memeory, 16 cores, and 8 32GB NVIDIA V100 GPU from Case Western Reserve University (CWRU) High Performance Computing (HPC) Cluster.
% Our source code, PV datasets, and a full version of the paper are made available \footnote{https://anonymous.4open.science/r/st-DynGNN-97EF}.  

\begin{table*}[tb!]
\caption{Comparison of Degradation Pattern Estimation Error between \stdyn and Baselines. The best-performing results are highlighted in bold, while the second-best ones are italicized. (``ST-GTrend-NF'': \stdyn w/o flatness regularization; ``ST-GTrend-NS'': \stdyn w/o smoothness regularization; ``-'': not available due to being PV domain methods).}
\vspace{-2ex}
\resizebox{\textwidth}{!}{%
\begin{tabular}{|l|cccccccccc|}
\hline
 &
  \multicolumn{10}{c|}{\textbf{Datasets}} \\ \cline{2-11} 
 &
  \multicolumn{2}{c|}{PV\_Case1} &
  \multicolumn{2}{c|}{PV\_Case2} &
  \multicolumn{2}{c|}{PV\_Case3} &
  \multicolumn{2}{c|}{Finance} &
  \multicolumn{2}{c|}{Economy} \\ \cline{2-11} 
\multirow{-3}{*}{\textbf{Models}} &
  \multicolumn{1}{c|}{MAPE} &
  \multicolumn{1}{c|}{ED} &
  \multicolumn{1}{c|}{MAPE} &
  \multicolumn{1}{c|}{ED} &
  \multicolumn{1}{c|}{MAPE} &
  \multicolumn{1}{c|}{ED} &
  \multicolumn{1}{c|}{MAPE} &
  \multicolumn{1}{c|}{ED} &
  \multicolumn{1}{c|}{MAPE} &
  ED \\ \hline
6K &
  \multicolumn{1}{c|}{11.97 ± 0.71} &
  \multicolumn{1}{c|}{0.410 ± 0.024} &
  \multicolumn{1}{c|}{11.11 ± 0.70} &
  \multicolumn{1}{c|}{0.424 ± 0.025} &
  \multicolumn{1}{c|}{12.14 ± 0.67} &
  \multicolumn{1}{c|}{0.410 ± 0.024} &
  \multicolumn{1}{c|}{-} &
  \multicolumn{1}{c|}{-} &
  \multicolumn{1}{c|}{-} &
  - \\ \hline
PVUSA &
  \multicolumn{1}{c|}{4.66 ± 0.26} &
  \multicolumn{1}{c|}{0.167 ± 0.010} &
  \multicolumn{1}{c|}{4.59 ± 0.26} &
  \multicolumn{1}{c|}{0.178 ± 0.010} &
  \multicolumn{1}{c|}{4.71 ± 0.25} &
  \multicolumn{1}{c|}{0.166 ± 0.010} &
  \multicolumn{1}{c|}{-} &
  \multicolumn{1}{c|}{-} &
  \multicolumn{1}{c|}{-} &
  - \\ \hline
XbX &
  \multicolumn{1}{c|}{4.21 ± 0.11} &
  \multicolumn{1}{c|}{0.147 ± 0.004} &
  \multicolumn{1}{c|}{4.20 ± 0.10} &
  \multicolumn{1}{c|}{0.161 ± 0.004} &
  \multicolumn{1}{c|}{4.18 ± 0.07} &
  \multicolumn{1}{c|}{0.144 ± 0.003} &
  \multicolumn{1}{c|}{-} &
  \multicolumn{1}{c|}{-} &
  \multicolumn{1}{c|}{-} &
  - \\ \hline
XbX+UTC &
  \multicolumn{1}{c|}{1.27 ± 0.04} &
  \multicolumn{1}{c|}{0.052 ± 0.002} &
  \multicolumn{1}{c|}{1.22 ± 0.04} &
  \multicolumn{1}{c|}{0.045 ± 0.002} &
  \multicolumn{1}{c|}{\textit{1.21 ± 0.04}} &
  \multicolumn{1}{c|}{\textit{0.044 ± 0.002}} &
  \multicolumn{1}{c|}{-} &
  \multicolumn{1}{c|}{-} &
  \multicolumn{1}{c|}{-} &
  - \\ \hline
QP-Aging-Detect &
  \multicolumn{1}{c|}{2.38 ± 0.21} &
  \multicolumn{1}{c|}{0.071 ± 0.005} &
  \multicolumn{1}{c|}{1.18 ± 0.07} &
  \multicolumn{1}{c|}{0.043 ± 0.002} &
  \multicolumn{1}{c|}{2.57 ± 0.24} &
  \multicolumn{1}{c|}{0.073 ± 0.005} &
  \multicolumn{1}{c|}{28.16 ± 1.52} &
  \multicolumn{1}{c|}{2.12  ± 0.12} &
  \multicolumn{1}{c|}{13.17 ± 0.97} &
  0.45 ± 0.02 \\ \hline
MSTL &
  \multicolumn{1}{c|}{1.84 ± 0.09} &
  \multicolumn{1}{c|}{0.055 ± 0.003} &
  \multicolumn{1}{c|}{1.90 ± 0.10} &
  \multicolumn{1}{c|}{0.064 ± 0.003} &
  \multicolumn{1}{c|}{1.45 ± 0.09} &
  \multicolumn{1}{c|}{0.041 ± 0.003} &
  \multicolumn{1}{c|}{24.47 ± 1.66} &
  \multicolumn{1}{c|}{1.94 ± 0.13} &
  \multicolumn{1}{c|}{10.70 ± 0.73} &
  0.37 ± 0.02 \\ \hline
SSA &
  \multicolumn{1}{c|}{2.03 ± 0.19} &
  \multicolumn{1}{c|}{0.059 ± 0.004} &
  \multicolumn{1}{c|}{1.13 ± 0.08} &
  \multicolumn{1}{c|}{0.041 ± 0.003} &
  \multicolumn{1}{c|}{1.96 ± 0.17} &
  \multicolumn{1}{c|}{0.054 ± 0.004} &
  \multicolumn{1}{c|}{21.72 ± 1.46} &
  \multicolumn{1}{c|}{1.70 ± 0.11} &
  \multicolumn{1}{c|}{17.37 ± 1.07} &
  0.57 ± 0.04 \\ \hline
STGAE1 &
  \multicolumn{1}{c|}{0.61 ± 0.03} &
  \multicolumn{1}{c|}{0.022 ± 0.002} &
  \multicolumn{1}{c|}{2.00 ± 0.18} &
  \multicolumn{1}{c|}{0.071 ± 0.006} &
  \multicolumn{1}{c|}{2.23 ± 0.19} &
  \multicolumn{1}{c|}{0.069 ± 0.005} &
  \multicolumn{1}{c|}{22.12 ± 1.33} &
  \multicolumn{1}{c|}{1.75 ± 0.11} &
  \multicolumn{1}{c|}{16.67 ± 1.02} &
  0.58 ± 0.03 \\ \hline
STGAE2 &
  \multicolumn{1}{c|}{\textit{0.46 ± 0.02}} &
  \multicolumn{1}{c|}{\textit{0.018 ± 0.001}} &
  \multicolumn{1}{c|}{2.04 ± 0.15} &
  \multicolumn{1}{c|}{0.075 ± 0.005} &
  \multicolumn{1}{c|}{1.81 ± 0.15} &
  \multicolumn{1}{c|}{0.053 ± 0.004} &
  \multicolumn{1}{c|}{21.85 ± 1.31} &
  \multicolumn{1}{c|}{1.72 ± 0.11} &
  \multicolumn{1}{c|}{10.53 ± 0.69} &
  0.36 ± 0.01 \\ \hline
ST-GTrend-NF &
  \multicolumn{1}{c|}{2.84 ± 0.19} &
  \multicolumn{1}{c|}{0.094 ± 0.005} &
  \multicolumn{1}{c|}{1.46 ± 0.11} &
  \multicolumn{1}{c|}{0.052 ± 0.003} &
  \multicolumn{1}{c|}{2.33 ± 0.14} &
  \multicolumn{1}{c|}{0.076 ± 0.005} &
  \multicolumn{1}{c|}{26.69 ± 1.48} &
  \multicolumn{1}{c|}{2.31 ± 0.13} &
  \multicolumn{1}{c|}{14.89 ± 0.86} &
  0.51 ± 0.01 \\ \hline
ST-GTrend-NS &
  \multicolumn{1}{c|}{0.69 ± 0.03} &
  \multicolumn{1}{c|}{0.027 ± 0.001} &
  \multicolumn{1}{c|}{\textit{1.01 ± 0.04}} &
  \multicolumn{1}{c|}{\textit{0.038 ± 0.001}} &
  \multicolumn{1}{c|}{1.95 ± 0.05} &
  \multicolumn{1}{c|}{0.068 ± 0.001} &
  \multicolumn{1}{c|}{\textit{21.26 ± 1.27}} &
  \multicolumn{1}{c|}{\textit{1.68 ± 0.10}} &
  \multicolumn{1}{c|}{\textit{9.49 ± 0.65}} &
  \textit{0.35 ± 0.01} \\ \hline
\textbf{ST-GTrend} &
  \multicolumn{1}{c|}{\textbf{0.40 ± 0.02}} &
  \multicolumn{1}{c|}{\textbf{0.015 ± 0.001}} &
  \multicolumn{1}{c|}{\textbf{0.96 ± 0.04}} &
  \multicolumn{1}{c|}{\textbf{0.036 ± 0.001}} &
  \multicolumn{1}{c|}{\textbf{0.85 ± 0.03}} &
  \multicolumn{1}{c|}{\textbf{0.031 ± 0.001}} &
  \multicolumn{1}{c|}{\textbf{18.65 ± 1.15}} &
  \multicolumn{1}{c|}{\textbf{1.44 ± 0.08}} &
  \multicolumn{1}{c|}{\textbf{9.31 ± 0.61}} &
   \textbf{0.31 ± 0.01} \\ \hline
\end{tabular}}%
\label{tab:Error_all}
\vspace{-5ex}
\end{table*}

\subsection{Experimental Results}

\stitle{Exp-1: PLR Degradation Estimation Errors}. 
To evaluate the PLR degradation pattern estimation accuracy, we compare MAPE and ED between the RDP and EDP.
Table~\ref{tab:Error_all} presents PLR degradation pattern estimation errors of \stdyn and three types of PV datasets, each exhibiting a different type of degradation pattern: linear, linear with breakpoint, and non-linear and two other datasets from finance and economy.

We observe that \stdyn achieves the best PLR degradation pattern estimation measured by MAPE and ED across all datasets.
Compared to the top two baselines XbX+UTC and STGAE2, \stdyn achieves a reduction of 34.74\% and 33.66\% on average in MAPE and ED. 
Moreover, even in the more challenging degradation patterns Case2 and Case3, \stdyn continues to outperform the other baselines significantly. 
These results demonstrate that \stdyn models have the potential to provide accurate estimation of PLR degradation patterns for all systems within a large PV fleet.

% Based on the scatter plots in Fig. \ref{fig:imputation_detail}, we find that the \stdyn model outperforms the mainstream imputation methods such as PVUSA and XbX in terms of capturing and imputing both the "up and down" (from cloudiness) and "plateau" (due to clipping) of PV power timeseries. 
% Specifically, the STGAE model produced more accurate predictions that closely followed the actual data points, while the predictions produced by MICE and KNN are more scattered and less accurate.
% These findings suggest that the proposed STGAE model is effective in capturing the complex spatial and temporal dependencies in PV datasets, which allows it to more accurately impute missing data points. 

\stitle{Exp-2: Detail Evaluations of \stdyn}. 
We conduct case studies to validate the
quality of disentaglement between aging and fluctuation terms decomposed by \stdyn and the effectiveness of the design of encoder an decoder in \stdyn. 
%We also showcase visualization of estimated degradation patterns derived from \stdyn.  %and top performed baselines compared to real degradation patterns.

\begin{figure}[tb!]
\vspace{-2ex}
% \begin{figure}
  \centering
  \includegraphics[width=0.95\linewidth]{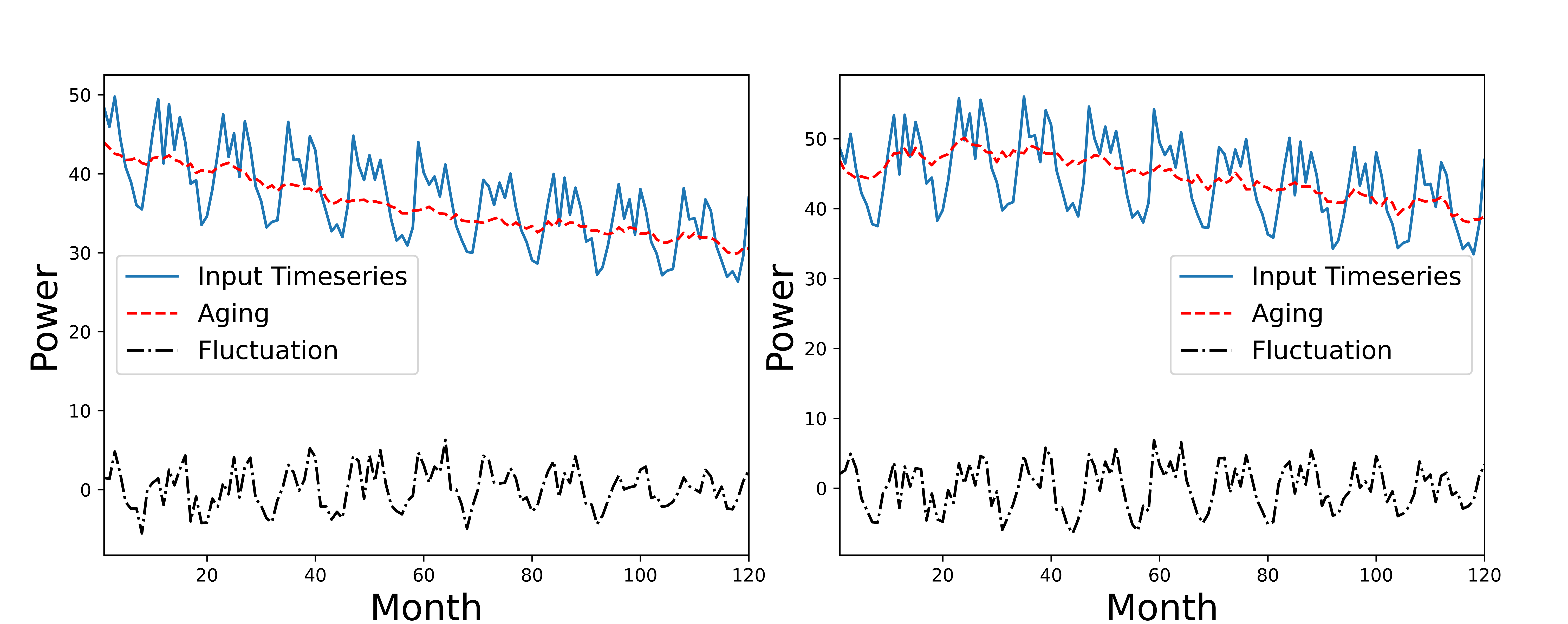}
   \vspace{-2ex}
  \caption{Example of Extracted Aging (in red) and Fluctuation Terms (in black) by \stdyn for a PV System (left: PV\_Case 3, right: PV\_Case 2).}
  \label{fig:decomposition_detail}
  \vspace{-4ex}
\end{figure}

\eetitle{Decomposition Results from \stdyn}. Fig. \ref{fig:decomposition_detail} illustrates extracted aging and fluctuation terms of a PV system under linear with breakpoint and non-linear degradation patterns.
From this figure, we observe that \stdyn achieves a clear disentanglement between aging and fluctuation terms.
Aging term successfully captures the initial upward trend in the cases 2 degradation pattern.
Fluctuation term captures annual seasonality and noises from the input timeseries.
Our results indicate that design of graph autoencoder and learning objective in \stdyn is effective in separating aging and fluctuation in PV dataset. 

\eetitle{Effectiveness of Encoder and Decoder Design}.
To further analyze the effectiveness of the encoder and decoder design of \stdyn, we quantify the estimation errors changes when replacing the graph convolution layers in \stdyn by either graph convolutional (GCN) layers in STGAE1 or graph attention networks (GAT) layers in STGAE2.
We observe that STGAE1 and STGAE2 incurs larger PLR estimation errors, on average 89.8\% and 65.7\% larger MAPE, when compared to \stdyn.
Our results verify the effectiveness of design of autoencoder in \stdyn.

\begin{figure*}[tb!]
% \vspace{-2ex}
  \centering
\includegraphics[width=0.95\linewidth]{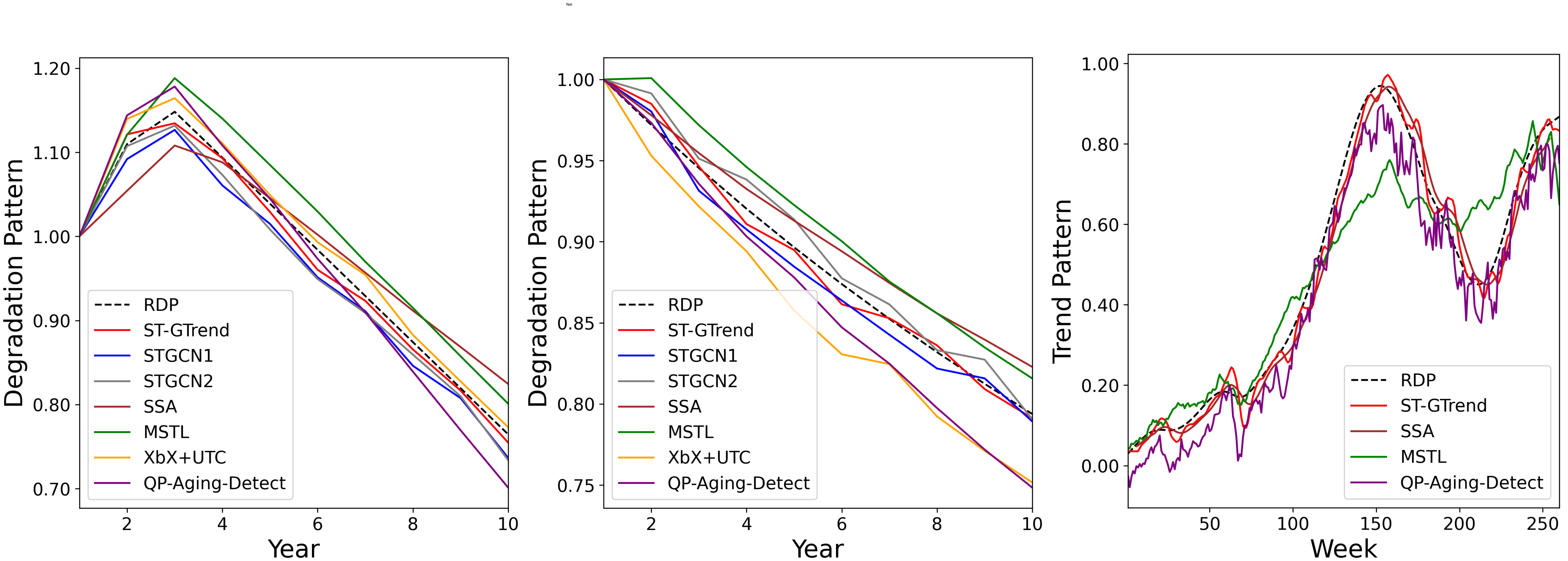}
\vspace{-2ex}
  \caption{Comparison of Real Degradation/Trend Pattern with Patterns Extracted by \stdyn and Baselines (left: PV\_Case 2; mid: PV\_Case 3; right: Finance).}
  \label{fig:visual_all}
  \vspace{-4ex}
\end{figure*}

\eetitle{Visual Analysis}. We compare EDP extracted by \stdyn and top six best-performed baselines with RDP.
Fig. \ref{fig:visual_all} shows \stdyn can better recover real degradation pattern from input timeseries than all baselines since it show a closer match with RDP. 
We can see EDP extracted by \stdyn is the closest to RDP in both case 2 and case 3 figures followed by XbX+UTC and STGAE2, which is consistent with the comparison of degradation pattern estimation error results shown in Table.~\ref{tab:Error_all}. 

\eat{\eetitle{Learning cost}. It is quite feasible to train \stdyn over 
large-scale PV fleet. For example, it takes on average $29.97$ seconds (averaged over 50 runs) to train 
\stdyn for a PV fleet consisting of 100 systems with 10-years timeseries.} 

% \begin{figure*}[tb!]
% % \vspace{-2ex}
%   \centering
% \includegraphics[width=0.9\linewidth]{fig_new/St-DynGNN_V1.png}
%   \caption{Overview of \stdyn Framework (six nodes shown in $G_{t}$ for illustration).}
%   \label{fig_new:Framework}
%   % \vspace{-2ex}
% \end{figure*}

\stitle{Exp-3: Ablation Analysis}.
%We conduct ablation analysis to verify the design of \stdyn, with its counterparts derived by removing certain components in the loss function.
To study how regularization terms imposed on learning objective impact the accuracy of \stdyn, 
we conduct two ablation analyses to remove the flatness regularization or smoothness regularization from our learning objective in Eq.~\ref{loss} to study how they affect the accuracy of \stdyn. 
As illustrated in Table.~\ref{tab:Error_all}, 
as for the flatness  regularization, we find that \stdyn with flatness  regularization reduces MAPE and ED by 58.10\% and 54.92\% on average compared to \stdyn without flatness regularization (ST-GTrend-NF). 
On the other hand, for smoothness regularization, we observe that \stdyn with smoothness regularization reduces MAPE and ED by 30.87\% and 31.27\% on average compared to \stdyn without smoothness regularization (ST-GTrend-NS). 
These results verify the effectiveness of regularization terms in the design of \stdyn.

\begin{figure}[tb!]
    \centering
    \begin{minipage}{0.22\textwidth}
        \centering
        \includegraphics[scale=0.5]{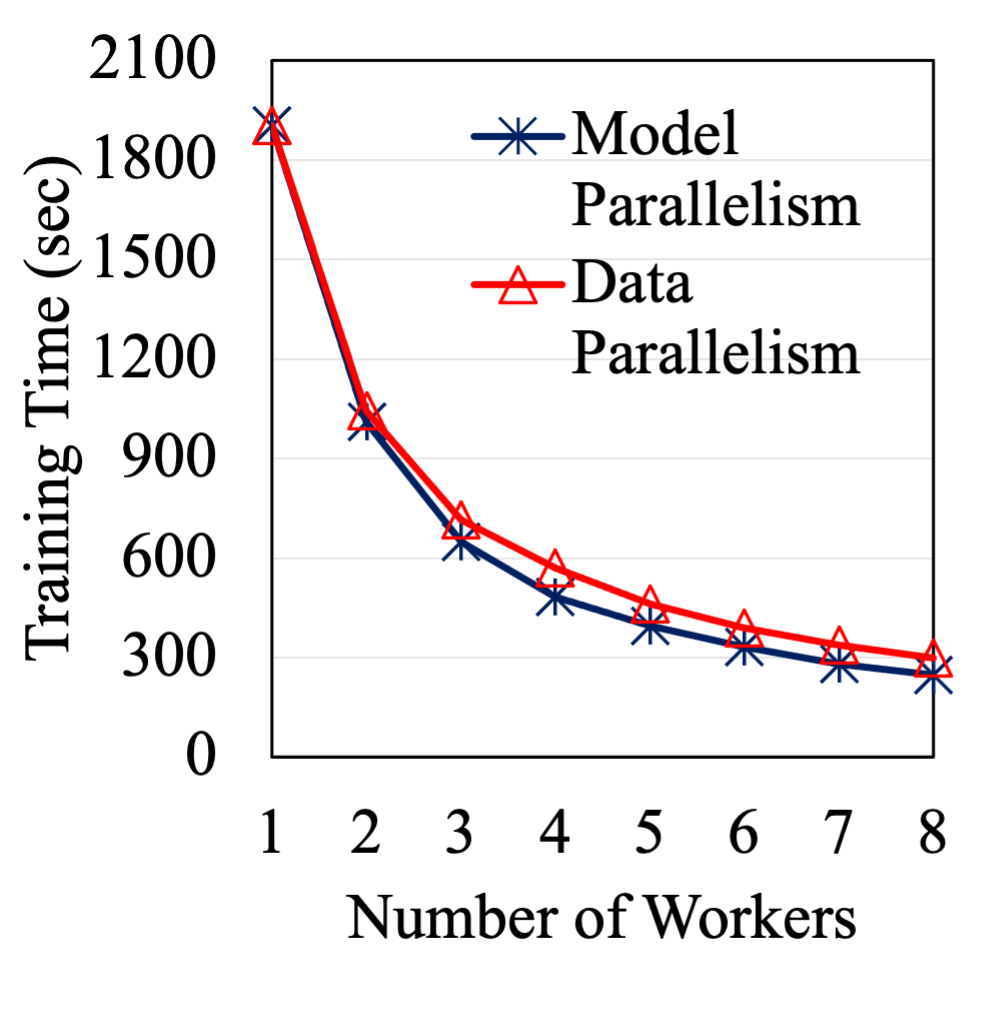}
        \vspace{-3ex}
        \subcaption{Dataset: PV}
    \end{minipage}
    % \hfill
    \begin{minipage}{0.22\textwidth}
        \centering
        \includegraphics[scale=0.5]{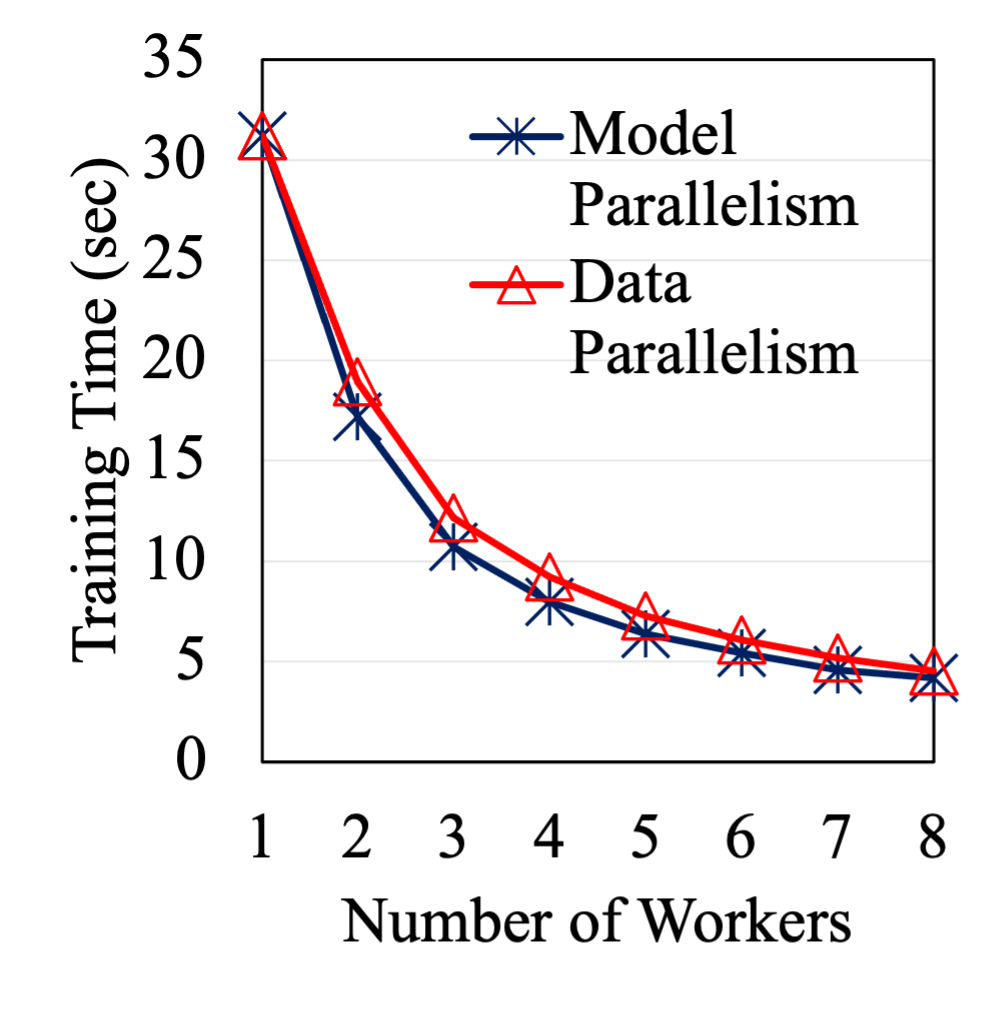}
        \vspace{-3ex}
        \subcaption{Dataset: Economy}
    \end{minipage}
    \vspace{-2ex}
    \caption{Scalability Test on Model and Data Parallelism (training time measured by averaging 50 rounds).}
    \label{fig:scalability}
    \vspace{-4ex}
\end{figure}

\stitle{Exp-4: Scalability Test}. 
We conduct controlled scalabilty tests on \stdyn %training by fixing either model or data to implement data or model parallelism and clock the training time acquired by either parallelism with varying number of workers.
to report the individual improvement of 
data or model parallelism alone. 
(1) We report the scalability of Level I (Model) Parallelism by distributing the training of GAE branches into parallel workers (CPUs). 
As shown in Fig.~\ref{fig:scalability}, the training time (on both PV and Economy datasets) is inverse proportional to the number of workers as it increases from 2 to 8. In other words, the speed-up achieved by model parallelism are proportional (linear) to the number of workers. 
(2) We next report the result for the Level II  (Data Parallelism) where we use multiple workers (GPUs) to train local GAE by fitting the batched input spatio-temporal graph snapshots in parallel.  %and backpropagate to update local gradients after each iteration.
Fig.~\ref{fig:scalability} verifies that \parastdyn scales well with more computing resources. The training efficiency of \stdyn 
reduces are improved by 7.56 times as the number of workers increases from 2 to 8.

\vspace{-2ex}
\section{Conclusion}\label{sec:conclusion}
We have proposed \stdyn, 
a novel spatiotemporal GNN that adopts paralleled graph autoencoder modules to decompose input photovoltaic (PV) power timeseries into aging (PV performance loss) and fluctuation terms for long-term trend (PLR) analysis. 
Each module exploits spatial coherence from neighboring PV inverters and temporal correlations within-series to perform aging and fluctuation extractions for PV fleet-level analysis. 
The loss function of \stdyn ensures the clear disentanglement between aging and fluctuation through smoothness and flatness regularizations. 
To accelerate \stdyn training and inference at scale, we have also developed \parastdyn, a hybrid parallelism algorithm, combining both model and data parallelism to facilitate scalable graph learning.
Our experiments have verified that \stdyn outperforms existing SOTA methods over PV datasets with diverse degradation patterns and datasets, generalizes to 
other types of datasets, and scales well 
with more computing resources. 
\eat{A future topic is to develop parallel training and inference of \stdyn for larger-scale PV systems.}

\eat{
\begin{acks}
This material is based upon work supported by the U.S. Department of Energy’s Office of Energy Efficiency and Renewable Energy (EERE) under Solar Energy Technologies Office (SETO) Agreement Number DE-EE0009353. 
%The authors acknowledge the support of the CWRU High Performance Computing.
\end{acks}
}

\bibliographystyle{ACM-Reference-Format}
\bibliography{references}


\begin{thebibliography}{52}

%%% the Web address.

\ifx \showCODEN    \undefined \def \showCODEN     #1{\unskip}     \fi
\ifx \showDOI      \undefined \def \showDOI       #1{#1}\fi
\ifx \showISBNx    \undefined \def \showISBNx     #1{\unskip}     \fi
\ifx \showISBNxiii \undefined \def \showISBNxiii  #1{\unskip}     \fi
\ifx \showISSN     \undefined \def \showISSN      #1{\unskip}     \fi
\ifx \showLCCN     \undefined \def \showLCCN      #1{\unskip}     \fi
\ifx \shownote     \undefined \def \shownote      #1{#1}          \fi
\ifx \showarticletitle \undefined \def \showarticletitle #1{#1}   \fi
\ifx \showURL      \undefined \def \showURL       {\relax}        \fi
% invisible to TeX
\providecommand\bibfield[2]{#2}
\providecommand\bibinfo[2]{#2}
\providecommand\natexlab[1]{#1}
\providecommand\showeprint[2][]{arXiv:#2}

\bibitem[Apaydin et~al\mbox{.}(2021)]%
        {apaydin2021artificial}
\bibfield{author}{\bibinfo{person}{Halit Apaydin}, \bibinfo{person}{Mohammad~Taghi Sattari}, \bibinfo{person}{Kambiz Falsafian}, {and} \bibinfo{person}{Ramendra Prasad}.} \bibinfo{year}{2021}\natexlab{}.
\newblock \showarticletitle{Artificial intelligence modelling integrated with Singular Spectral analysis and Seasonal-Trend decomposition using Loess approaches for streamflow predictions}.
\newblock \bibinfo{journal}{\emph{Journal of Hydrology}}  \bibinfo{volume}{600} (\bibinfo{year}{2021}), \bibinfo{pages}{126506}.
\newblock


\bibitem[Bai et~al\mbox{.}(2020)]%
        {bai2020adaptive}
\bibfield{author}{\bibinfo{person}{Lei Bai}, \bibinfo{person}{Lina Yao}, \bibinfo{person}{Can Li}, \bibinfo{person}{Xianzhi Wang}, {and} \bibinfo{person}{Can Wang}.} \bibinfo{year}{2020}\natexlab{}.
\newblock \showarticletitle{Adaptive graph convolutional recurrent network for traffic forecasting}.
\newblock \bibinfo{journal}{\emph{Advances in Neural Information Processing Systems}}  \bibinfo{volume}{33} (\bibinfo{year}{2020}), \bibinfo{pages}{17804--17815}.
\newblock


\bibitem[Ben-Nun and Hoefler(2019)]%
        {ben2019demystifying}
\bibfield{author}{\bibinfo{person}{Tal Ben-Nun} {and} \bibinfo{person}{Torsten Hoefler}.} \bibinfo{year}{2019}\natexlab{}.
\newblock \showarticletitle{Demystifying parallel and distributed deep learning: An in-depth concurrency analysis}.
\newblock \bibinfo{journal}{\emph{ACM Computing Surveys (CSUR)}} \bibinfo{volume}{52}, \bibinfo{number}{4} (\bibinfo{year}{2019}), \bibinfo{pages}{1--43}.
\newblock


\bibitem[Besta and Hoefler(2022)]%
        {besta2022parallel}
\bibfield{author}{\bibinfo{person}{Maciej Besta} {and} \bibinfo{person}{Torsten Hoefler}.} \bibinfo{year}{2022}\natexlab{}.
\newblock \showarticletitle{Parallel and distributed graph neural networks: An in-depth concurrency analysis}.
\newblock \bibinfo{journal}{\emph{arXiv preprint arXiv:2205.09702}} (\bibinfo{year}{2022}).
\newblock


\bibitem[Curran et~al\mbox{.}(2017)]%
        {curran2017determining}
\bibfield{author}{\bibinfo{person}{Alan~J Curran}, \bibinfo{person}{Yang Hu}, \bibinfo{person}{Rojiar Haddadian}, \bibinfo{person}{Jennifer~L Braid}, \bibinfo{person}{David Meakin}, \bibinfo{person}{Timothy~J Peshek}, {and} \bibinfo{person}{Roger~H French}.} \bibinfo{year}{2017}\natexlab{}.
\newblock \showarticletitle{Determining the power rate of change of 353 plant inverters time-series data across multiple climate zones, using a month-by-month data science analysis}. In \bibinfo{booktitle}{\emph{2017 IEEE 44th Photovoltaic Specialist Conference (PVSC)}}. \bibinfo{publisher}{IEEE}, \bibinfo{pages}{1927--1932}.
\newblock


\bibitem[Dai and Zhu(2020)]%
        {dai2020forecasting}
\bibfield{author}{\bibinfo{person}{Zhifeng Dai} {and} \bibinfo{person}{Huan Zhu}.} \bibinfo{year}{2020}\natexlab{}.
\newblock \showarticletitle{Forecasting stock market returns by combining sum-of-the-parts and ensemble empirical mode decomposition}.
\newblock \bibinfo{journal}{\emph{Applied Economics}} \bibinfo{volume}{52}, \bibinfo{number}{21} (\bibinfo{year}{2020}), \bibinfo{pages}{2309--2323}.
\newblock


\bibitem[Darling et~al\mbox{.}(2011)]%
        {darling2011assumptions}
\bibfield{author}{\bibinfo{person}{Seth~B Darling}, \bibinfo{person}{Fengqi You}, \bibinfo{person}{Thomas Veselka}, {and} \bibinfo{person}{Alfonso Velosa}.} \bibinfo{year}{2011}\natexlab{}.
\newblock \showarticletitle{Assumptions and the levelized cost of energy for photovoltaics}.
\newblock \bibinfo{journal}{\emph{Energy \& environmental science}} \bibinfo{volume}{4}, \bibinfo{number}{9} (\bibinfo{year}{2011}), \bibinfo{pages}{3133--3139}.
\newblock


\bibitem[Deline et~al\mbox{.}(2022)]%
        {delineReducingUncertaintyFielded2022b}
\bibfield{author}{\bibinfo{person}{Chris Deline}, \bibinfo{person}{Mike Deceglie}, \bibinfo{person}{Dirk Jordan}, \bibinfo{person}{Matt Muller}, \bibinfo{person}{Kevin Anderson}, \bibinfo{person}{Kirsten Perry}, \bibinfo{person}{Robert White}, {and} \bibinfo{person}{Mark Bolinger}.} \bibinfo{year}{2022}\natexlab{}.
\newblock \bibinfo{booktitle}{\emph{Reducing {{Uncertainty}} of {{Fielded Photovoltaic Performance}} ({{Final Technical Report}})}}.
\newblock \bibinfo{type}{{T}echnical {R}eport} NREL/TP-5K00-82816. \bibinfo{institution}{{National Renewable Energy Lab. (NREL), Golden, CO (United States)}}.
\newblock
\urldef\tempurl%
\url{https://doi.org/10.2172/1880076}
\showDOI{\tempurl}


\bibitem[Deng et~al\mbox{.}(2022)]%
        {deng2022multi}
\bibfield{author}{\bibinfo{person}{Changrui Deng}, \bibinfo{person}{Yanmei Huang}, \bibinfo{person}{Najmul Hasan}, {and} \bibinfo{person}{Yukun Bao}.} \bibinfo{year}{2022}\natexlab{}.
\newblock \showarticletitle{Multi-step-ahead stock price index forecasting using long short-term memory model with multivariate empirical mode decomposition}.
\newblock \bibinfo{journal}{\emph{Information Sciences}}  \bibinfo{volume}{607} (\bibinfo{year}{2022}), \bibinfo{pages}{297--321}.
\newblock


\bibitem[Diamond and Boyd(2016)]%
        {diamond2016cvxpy}
\bibfield{author}{\bibinfo{person}{Steven Diamond} {and} \bibinfo{person}{Stephen Boyd}.} \bibinfo{year}{2016}\natexlab{}.
\newblock \showarticletitle{{CVXPY}: {A} {P}ython-embedded modeling language for convex optimization}.
\newblock \bibinfo{journal}{\emph{Journal of Machine Learning Research}} \bibinfo{volume}{17}, \bibinfo{number}{83} (\bibinfo{year}{2016}), \bibinfo{pages}{1--5}.
\newblock


\bibitem[Fan et~al\mbox{.}(2014)]%
        {fan2014distributed}
\bibfield{author}{\bibinfo{person}{Wenfei Fan}, \bibinfo{person}{Xin Wang}, \bibinfo{person}{Yinghui Wu}, {and} \bibinfo{person}{Dong Deng}.} \bibinfo{year}{2014}\natexlab{}.
\newblock \showarticletitle{Distributed graph simulation: Impossibility and possibility}.
\newblock \bibinfo{journal}{\emph{Proceedings of the VLDB Endowment}} \bibinfo{volume}{7}, \bibinfo{number}{12} (\bibinfo{year}{2014}), \bibinfo{pages}{1083--1094}.
\newblock


\bibitem[Fan et~al\mbox{.}(2023)]%
        {10.1145/3588730}
\bibfield{author}{\bibinfo{person}{Yangxin Fan}, \bibinfo{person}{Xuanji Yu}, \bibinfo{person}{Raymond Wieser}, \bibinfo{person}{David Meakin}, \bibinfo{person}{Avishai Shaton}, \bibinfo{person}{Jean-Nicolas Jaubert}, \bibinfo{person}{Robert Flottemesch}, \bibinfo{person}{Michael Howell}, \bibinfo{person}{Jennifer Braid}, \bibinfo{person}{Laura Bruckman}, \bibinfo{person}{Roger French}, {and} \bibinfo{person}{Yinghui Wu}.} \bibinfo{year}{2023}\natexlab{}.
\newblock \showarticletitle{Spatio-Temporal Denoising Graph Autoencoders with Data Augmentation for Photovoltaic Data Imputation}.
\newblock \bibinfo{journal}{\emph{Proc. ACM Manag. Data}} \bibinfo{volume}{1}, \bibinfo{number}{1}, Article \bibinfo{articleno}{50} (\bibinfo{date}{may} \bibinfo{year}{2023}), \bibinfo{numpages}{19}~pages.
\newblock
\urldef\tempurl%
\url{https://doi.org/10.1145/3588730}
\showDOI{\tempurl}


\bibitem[Fan et~al\mbox{.}(2024)]%
        {PVplr2024}
\bibfield{author}{\bibinfo{person}{Yangxin Fan}, \bibinfo{person}{Xuanji Yu}, \bibinfo{person}{Raymond Wieser}, \bibinfo{person}{Yinghui Wu}, {and} \bibinfo{person}{Roger French}.} \bibinfo{year}{2024}\natexlab{}.
\newblock \bibinfo{booktitle}{\emph{PVplr-stGNN}}.
\newblock SDLE Research Center, Case Western Reserve Universitu.
\newblock
\urldef\tempurl%
\url{https://pypi.org/project/PVplr-stGNN/}
\showURL{%
\tempurl}
\newblock
\shownote{PyPI}.


\bibitem[Feng et~al\mbox{.}(2021)]%
        {feng2021context}
\bibfield{author}{\bibinfo{person}{Jie Feng}, \bibinfo{person}{Yong Li}, \bibinfo{person}{Ziqian Lin}, \bibinfo{person}{Can Rong}, \bibinfo{person}{Funing Sun}, \bibinfo{person}{Diansheng Guo}, {and} \bibinfo{person}{Depeng Jin}.} \bibinfo{year}{2021}\natexlab{}.
\newblock \showarticletitle{Context-aware spatial-temporal neural network for citywide crowd flow prediction via modeling long-range spatial dependency}.
\newblock \bibinfo{journal}{\emph{ACM Transactions on Knowledge Discovery from Data (TKDD)}} \bibinfo{volume}{16}, \bibinfo{number}{3} (\bibinfo{year}{2021}), \bibinfo{pages}{1--21}.
\newblock


\bibitem[Fey and Lenssen(2019)]%
        {Fey/Lenssen/2019}
\bibfield{author}{\bibinfo{person}{Matthias Fey} {and} \bibinfo{person}{Jan~E. Lenssen}.} \bibinfo{year}{2019}\natexlab{}.
\newblock \showarticletitle{Fast Graph Representation Learning with {PyTorch Geometric}}. In \bibinfo{booktitle}{\emph{ICLR Workshop on Representation Learning on Graphs and Manifolds}}.
\newblock


\bibitem[French et~al\mbox{.}(2021)]%
        {french2021assessment}
\bibfield{author}{\bibinfo{person}{Roger~H French}, \bibinfo{person}{Laura~S Bruckman}, \bibinfo{person}{David Moser}, \bibinfo{person}{EURAC Lindig}, \bibinfo{person}{Mike van Iseghem}, \bibinfo{person}{Joshua~S Stein}, \bibinfo{person}{Maurice Richter}, \bibinfo{person}{Magnus Herz}, \bibinfo{person}{Wilfried van Sark}, \bibinfo{person}{Franz Baumgartner}, {et~al\mbox{.}}} \bibinfo{year}{2021}\natexlab{}.
\newblock \showarticletitle{Assessment of performance loss rate of PV power systems}.
\newblock  (\bibinfo{year}{2021}).
\newblock


\bibitem[Gao et~al\mbox{.}(2022)]%
        {gao2022efficient}
\bibfield{author}{\bibinfo{person}{Xinyi Gao}, \bibinfo{person}{Wentao Zhang}, \bibinfo{person}{Yingxia Shao}, \bibinfo{person}{Quoc Viet~Hung Nguyen}, \bibinfo{person}{Bin Cui}, {and} \bibinfo{person}{Hongzhi Yin}.} \bibinfo{year}{2022}\natexlab{}.
\newblock \showarticletitle{Efficient Graph Neural Network Inference at Large Scale}.
\newblock \bibinfo{journal}{\emph{arXiv preprint arXiv:2211.00495}} (\bibinfo{year}{2022}).
\newblock


\bibitem[Gazbour et~al\mbox{.}(2018)]%
        {gazbour2018path}
\bibfield{author}{\bibinfo{person}{Nouha Gazbour}, \bibinfo{person}{Guillaume Razongles}, \bibinfo{person}{Elise Monnier}, \bibinfo{person}{Maryline Joanny}, \bibinfo{person}{Carole Charbuillet}, \bibinfo{person}{Fran{\c{c}}oise Burgun}, {and} \bibinfo{person}{Christian Schaeffer}.} \bibinfo{year}{2018}\natexlab{}.
\newblock \showarticletitle{A path to reduce variability of the environmental footprint results of photovoltaic systems}.
\newblock \bibinfo{journal}{\emph{Journal of cleaner production}}  \bibinfo{volume}{197} (\bibinfo{year}{2018}), \bibinfo{pages}{1607--1618}.
\newblock


\bibitem[Gorjian et~al\mbox{.}(2010)]%
        {gorjian2010review}
\bibfield{author}{\bibinfo{person}{Nima Gorjian}, \bibinfo{person}{Lin Ma}, \bibinfo{person}{Murthy Mittinty}, \bibinfo{person}{Prasad Yarlagadda}, {and} \bibinfo{person}{Yong Sun}.} \bibinfo{year}{2010}\natexlab{}.
\newblock \showarticletitle{A review on degradation models in reliability analysis}. In \bibinfo{booktitle}{\emph{Engineering Asset Lifecycle Management: Proceedings of the 4th World Congress on Engineering Asset Management (WCEAM 2009), 28-30 September 2009}}. \bibinfo{publisher}{Springer}, \bibinfo{pages}{369--384}.
\newblock


\bibitem[Hegedus et~al\mbox{.}(2000)]%
        {hegedus2000analysis}
\bibfield{author}{\bibinfo{person}{Steven~S Hegedus}, \bibinfo{person}{Brian~E McCandless}, {and} \bibinfo{person}{Robert~W Birkmire}.} \bibinfo{year}{2000}\natexlab{}.
\newblock \showarticletitle{Analysis of stress-induced degradation in CdS/CdTe solar cells}. In \bibinfo{booktitle}{\emph{Conference Record of the Twenty-Eighth IEEE Photovoltaic Specialists Conference-2000 (Cat. No. 00CH37036)}}. IEEE, \bibinfo{pages}{535--538}.
\newblock


\bibitem[Huang et~al\mbox{.}(1998)]%
        {huang1998empirical}
\bibfield{author}{\bibinfo{person}{Norden~E Huang}, \bibinfo{person}{Zheng Shen}, \bibinfo{person}{Steven~R Long}, \bibinfo{person}{Manli~C Wu}, \bibinfo{person}{Hsing~H Shih}, \bibinfo{person}{Quanan Zheng}, \bibinfo{person}{Nai-Chyuan Yen}, \bibinfo{person}{Chi~Chao Tung}, {and} \bibinfo{person}{Henry~H Liu}.} \bibinfo{year}{1998}\natexlab{}.
\newblock \showarticletitle{The empirical mode decomposition and the Hilbert spectrum for nonlinear and non-stationary time series analysis}.
\newblock \bibinfo{journal}{\emph{Proceedings of the Royal Society of London. Series A: mathematical, physical and engineering sciences}} \bibinfo{volume}{454}, \bibinfo{number}{1971} (\bibinfo{year}{1998}), \bibinfo{pages}{903--995}.
\newblock


\bibitem[Huld et~al\mbox{.}(2011)]%
        {huld2011power}
\bibfield{author}{\bibinfo{person}{Thomas Huld}, \bibinfo{person}{Gabi Friesen}, \bibinfo{person}{Artur Skoczek}, \bibinfo{person}{Robert~P Kenny}, \bibinfo{person}{Tony Sample}, \bibinfo{person}{Michael Field}, {and} \bibinfo{person}{Ewan~D Dunlop}.} \bibinfo{year}{2011}\natexlab{}.
\newblock \showarticletitle{A power-rating model for crystalline silicon PV modules}.
\newblock \bibinfo{journal}{\emph{Solar Energy Materials and Solar Cells}} \bibinfo{volume}{95}, \bibinfo{number}{12} (\bibinfo{year}{2011}), \bibinfo{pages}{3359--3369}.
\newblock


\bibitem[Ingenhoven et~al\mbox{.}(2017)]%
        {ingenhoven2017comparison}
\bibfield{author}{\bibinfo{person}{Philip Ingenhoven}, \bibinfo{person}{Giorgio Belluardo}, {and} \bibinfo{person}{David Moser}.} \bibinfo{year}{2017}\natexlab{}.
\newblock \showarticletitle{Comparison of statistical and deterministic smoothing methods to reduce the uncertainty of performance loss rate estimates}.
\newblock \bibinfo{journal}{\emph{IEEE Journal of Photovoltaics}} \bibinfo{volume}{8}, \bibinfo{number}{1} (\bibinfo{year}{2017}), \bibinfo{pages}{224--232}.
\newblock


\bibitem[Jordan et~al\mbox{.}(2017a)]%
        {jordan2017robust}
\bibfield{author}{\bibinfo{person}{Dirk~C Jordan}, \bibinfo{person}{Chris Deline}, \bibinfo{person}{Sarah~R Kurtz}, \bibinfo{person}{Gregory~M Kimball}, {and} \bibinfo{person}{Mike Anderson}.} \bibinfo{year}{2017}\natexlab{a}.
\newblock \showarticletitle{Robust PV degradation methodology and application}.
\newblock \bibinfo{journal}{\emph{IEEE Journal of photovoltaics}} \bibinfo{volume}{8}, \bibinfo{number}{2} (\bibinfo{year}{2017}), \bibinfo{pages}{525--531}.
\newblock


\bibitem[Jordan et~al\mbox{.}(2017b)]%
        {jordanPVDegradationCurves2017}
\bibfield{author}{\bibinfo{person}{Dirk~C. Jordan}, \bibinfo{person}{Timothy~J. Silverman}, \bibinfo{person}{Bill Sekulic}, {and} \bibinfo{person}{Sarah~R. Kurtz}.} \bibinfo{year}{2017}\natexlab{b}.
\newblock \showarticletitle{{{PV}} Degradation Curves: Non-Linearities and Failure Modes}.
\newblock \bibinfo{journal}{\emph{Progress in Photovoltaics: Research and Applications}} \bibinfo{volume}{25}, \bibinfo{number}{7} (\bibinfo{date}{July} \bibinfo{year}{2017}), \bibinfo{pages}{583--591}.
\newblock
\showISSN{1099-159X}
\urldef\tempurl%
\url{https://doi.org/10.1002/pip.2835}
\showDOI{\tempurl}


\bibitem[Kaaya et~al\mbox{.}(2020)]%
        {kaaya2020photovoltaic}
\bibfield{author}{\bibinfo{person}{Ismail Kaaya}, \bibinfo{person}{Sascha Lindig}, \bibinfo{person}{Karl-Anders Weiss}, \bibinfo{person}{Alessandro Virtuani}, \bibinfo{person}{Mariano Sidrach~de Cardona~Ortin}, {and} \bibinfo{person}{David Moser}.} \bibinfo{year}{2020}\natexlab{}.
\newblock \showarticletitle{Photovoltaic lifetime forecast model based on degradation patterns}.
\newblock \bibinfo{journal}{\emph{Progress in Photovoltaics: Research and Applications}} \bibinfo{volume}{28}, \bibinfo{number}{10} (\bibinfo{year}{2020}), \bibinfo{pages}{979--992}.
\newblock


\bibitem[Karimi et~al\mbox{.}(2021)]%
        {karimi2021spatiotemporal}
\bibfield{author}{\bibinfo{person}{Ahmad~Maroof Karimi}, \bibinfo{person}{Yinghui Wu}, \bibinfo{person}{Mehmet Koyuturk}, {and} \bibinfo{person}{Roger~H French}.} \bibinfo{year}{2021}\natexlab{}.
\newblock \showarticletitle{Spatiotemporal Graph Neural Network for Performance Prediction of Photovoltaic Power Systems}. In \bibinfo{booktitle}{\emph{Proceedings of the AAAI Conference on Artificial Intelligence}}.
\newblock


\bibitem[King et~al\mbox{.}(1997)]%
        {king1997field}
\bibfield{author}{\bibinfo{person}{David~L King}, \bibinfo{person}{Jay~A Kratochvil}, {and} \bibinfo{person}{William~E Boyson}.} \bibinfo{year}{1997}\natexlab{}.
\newblock \showarticletitle{Field experience with a new performance characterization procedure for photovoltaic arrays}.
\newblock \bibinfo{journal}{\emph{Sandia National Lab.(SNL-NM), Albuquerque, NM (United States)}} (\bibinfo{year}{1997}).
\newblock


\bibitem[Kipf and Welling(2017)]%
        {kipf2016semi}
\bibfield{author}{\bibinfo{person}{Thomas~N. Kipf} {and} \bibinfo{person}{Max Welling}.} \bibinfo{year}{2017}\natexlab{}.
\newblock \showarticletitle{Semi-Supervised Classification with Graph Convolutional Networks}. In \bibinfo{booktitle}{\emph{ICLR}}.
\newblock


\bibitem[Kumar and Gupta(1991)]%
        {kumar1991analysis}
\bibfield{author}{\bibinfo{person}{Vipin Kumar} {and} \bibinfo{person}{Anshul Gupta}.} \bibinfo{year}{1991}\natexlab{}.
\newblock \showarticletitle{Analysis of scalability of parallel algorithms and architectures: A survey}. In \bibinfo{booktitle}{\emph{Proceedings of the 5th international conference on Supercomputing}}. \bibinfo{pages}{396--405}.
\newblock


\bibitem[Lindig et~al\mbox{.}(2018)]%
        {lindig2018review}
\bibfield{author}{\bibinfo{person}{Sascha Lindig}, \bibinfo{person}{Ismail Kaaya}, \bibinfo{person}{Karl-Anders Wei{\ss}}, \bibinfo{person}{David Moser}, {and} \bibinfo{person}{Marko Topic}.} \bibinfo{year}{2018}\natexlab{}.
\newblock \showarticletitle{Review of statistical and analytical degradation models for photovoltaic modules and systems as well as related improvements}.
\newblock \bibinfo{journal}{\emph{IEEE Journal of Photovoltaics}} \bibinfo{volume}{8}, \bibinfo{number}{6} (\bibinfo{year}{2018}), \bibinfo{pages}{1773--1786}.
\newblock


\bibitem[Lindig et~al\mbox{.}(2022)]%
        {lindig2022best}
\bibfield{author}{\bibinfo{person}{Sascha Lindig}, \bibinfo{person}{Marios Theristis}, {and} \bibinfo{person}{David Moser}.} \bibinfo{year}{2022}\natexlab{}.
\newblock \showarticletitle{Best practices for photovoltaic performance loss rate calculations}.
\newblock \bibinfo{journal}{\emph{Progress in Energy}} \bibinfo{volume}{4}, \bibinfo{number}{2} (\bibinfo{year}{2022}), \bibinfo{pages}{022003}.
\newblock


\bibitem[Makrides et~al\mbox{.}(2014)]%
        {makrides2014performance}
\bibfield{author}{\bibinfo{person}{George Makrides}, \bibinfo{person}{Bastian Zinsser}, \bibinfo{person}{Markus Schubert}, {and} \bibinfo{person}{George~E Georghiou}.} \bibinfo{year}{2014}\natexlab{}.
\newblock \showarticletitle{Performance loss rate of twelve photovoltaic technologies under field conditions using statistical techniques}.
\newblock \bibinfo{journal}{\emph{Solar Energy}}  \bibinfo{volume}{103} (\bibinfo{year}{2014}), \bibinfo{pages}{28--42}.
\newblock


\bibitem[Mao et~al\mbox{.}(2020)]%
        {mao2020analysis}
\bibfield{author}{\bibinfo{person}{Xuegeng Mao}, \bibinfo{person}{Albert~C Yang}, \bibinfo{person}{Chung-Kang Peng}, {and} \bibinfo{person}{Pengjian Shang}.} \bibinfo{year}{2020}\natexlab{}.
\newblock \showarticletitle{Analysis of economic growth fluctuations based on EEMD and causal decomposition}.
\newblock \bibinfo{journal}{\emph{Physica A: Statistical Mechanics and its Applications}}  \bibinfo{volume}{553} (\bibinfo{year}{2020}), \bibinfo{pages}{124661}.
\newblock


\bibitem[Meyers(2022)]%
        {meyersEstimationSoilingLosses2022}
\bibfield{author}{\bibinfo{person}{Bennet Meyers}.} \bibinfo{year}{2022}\natexlab{}.
\newblock \showarticletitle{Estimation of {{Soiling Losses}} in {{Unlabeled PV Data}}}. In \bibinfo{booktitle}{\emph{2022 {{IEEE}} 49th {{Photovoltaics Specialists Conference}} ({{PVSC}})}}. \bibinfo{pages}{0930--0936}.
\newblock
\urldef\tempurl%
\url{https://doi.org/10.1109/PVSC48317.2022.9938567}
\showDOI{\tempurl}


\bibitem[Narayanan et~al\mbox{.}(2019)]%
        {narayanan2019pipedream}
\bibfield{author}{\bibinfo{person}{Deepak Narayanan}, \bibinfo{person}{Aaron Harlap}, \bibinfo{person}{Amar Phanishayee}, \bibinfo{person}{Vivek Seshadri}, \bibinfo{person}{Nikhil~R Devanur}, \bibinfo{person}{Gregory~R Ganger}, \bibinfo{person}{Phillip~B Gibbons}, {and} \bibinfo{person}{Matei Zaharia}.} \bibinfo{year}{2019}\natexlab{}.
\newblock \showarticletitle{PipeDream: Generalized pipeline parallelism for DNN training}. In \bibinfo{booktitle}{\emph{Proceedings of the 27th ACM Symposium on Operating Systems Principles}}. \bibinfo{pages}{1--15}.
\newblock


\bibitem[Nihar et~al\mbox{.}(2021)]%
        {nihar2021toward}
\bibfield{author}{\bibinfo{person}{Arafath Nihar}, \bibinfo{person}{Alan~J Curran}, \bibinfo{person}{Ahmad~M Karimi}, \bibinfo{person}{Jennifer~L Braid}, \bibinfo{person}{Laura~S Bruckman}, \bibinfo{person}{Mehmet Koyut{\"u}rk}, \bibinfo{person}{Yinghui Wu}, {and} \bibinfo{person}{Roger~H French}.} \bibinfo{year}{2021}\natexlab{}.
\newblock \showarticletitle{Toward findable, accessible, interoperable and reusable (fair) photovoltaic system time series data}. In \bibinfo{booktitle}{\emph{2021 IEEE 48th Photovoltaic Specialists Conference (PVSC)}}. \bibinfo{publisher}{IEEE}, \bibinfo{pages}{1701--1706}.
\newblock


\bibitem[Pontes et~al\mbox{.}(2016)]%
        {pontes2016design}
\bibfield{author}{\bibinfo{person}{Fabr{\'\i}cio~Jos{\'e} Pontes}, \bibinfo{person}{GF Amorim}, \bibinfo{person}{Pedro~Paulo Balestrassi}, \bibinfo{person}{AP Paiva}, {and} \bibinfo{person}{Jo{\~a}o~Roberto Ferreira}.} \bibinfo{year}{2016}\natexlab{}.
\newblock \showarticletitle{Design of experiments and focused grid search for neural network parameter optimization}.
\newblock \bibinfo{journal}{\emph{Neurocomputing}}  \bibinfo{volume}{186} (\bibinfo{year}{2016}), \bibinfo{pages}{22--34}.
\newblock


\bibitem[Rozemberczki et~al\mbox{.}(2021)]%
        {rozemberczki2021pytorch}
\bibfield{author}{\bibinfo{person}{Benedek Rozemberczki}, \bibinfo{person}{Paul Scherer}, \bibinfo{person}{Yixuan He}, \bibinfo{person}{George Panagopoulos}, \bibinfo{person}{Alexander Riedel}, \bibinfo{person}{Maria Astefanoaei}, \bibinfo{person}{Oliver Kiss}, \bibinfo{person}{Ferenc Beres}, \bibinfo{person}{Guzman Lopez}, \bibinfo{person}{Nicolas Collignon}, {and} \bibinfo{person}{Rik Sarkar}.} \bibinfo{year}{2021}\natexlab{}.
\newblock \showarticletitle{{PyTorch Geometric Temporal: Spatiotemporal Signal Processing with Neural Machine Learning Models}}. In \bibinfo{booktitle}{\emph{Proceedings of the 30th ACM International Conference on Information and Knowledge Management}}.
\newblock


\bibitem[Sa{\^a}daoui and Messaoud(2020)]%
        {saadaoui2020multiscaled}
\bibfield{author}{\bibinfo{person}{Foued Sa{\^a}daoui} {and} \bibinfo{person}{Othman~Ben Messaoud}.} \bibinfo{year}{2020}\natexlab{}.
\newblock \showarticletitle{Multiscaled neural autoregressive distributed lag: A new empirical mode decomposition model for nonlinear time series forecasting}.
\newblock \bibinfo{journal}{\emph{International Journal of Neural Systems}} \bibinfo{volume}{30}, \bibinfo{number}{08} (\bibinfo{year}{2020}), \bibinfo{pages}{2050039}.
\newblock


\bibitem[Seo et~al\mbox{.}(2018)]%
        {seo2018structured}
\bibfield{author}{\bibinfo{person}{Youngjoo Seo}, \bibinfo{person}{Micha{\"e}l Defferrard}, \bibinfo{person}{Pierre Vandergheynst}, {and} \bibinfo{person}{Xavier Bresson}.} \bibinfo{year}{2018}\natexlab{}.
\newblock \showarticletitle{Structured sequence modeling with graph convolutional recurrent networks}. In \bibinfo{booktitle}{\emph{International Conference on Neural Information Processing}}. \bibinfo{publisher}{Springer}, \bibinfo{pages}{362--373}.
\newblock


\bibitem[Shahraki et~al\mbox{.}(2017)]%
        {shahraki2017review}
\bibfield{author}{\bibinfo{person}{Ameneh~Forouzandeh Shahraki}, \bibinfo{person}{Om~Parkash Yadav}, {and} \bibinfo{person}{Haitao Liao}.} \bibinfo{year}{2017}\natexlab{}.
\newblock \showarticletitle{A review on degradation modelling and its engineering applications}.
\newblock \bibinfo{journal}{\emph{International Journal of Performability Engineering}} \bibinfo{volume}{13}, \bibinfo{number}{3} (\bibinfo{year}{2017}), \bibinfo{pages}{299}.
\newblock


\bibitem[Shi et~al\mbox{.}(2020)]%
        {shi2020masked}
\bibfield{author}{\bibinfo{person}{Yunsheng Shi}, \bibinfo{person}{Zhengjie Huang}, \bibinfo{person}{Shikun Feng}, \bibinfo{person}{Hui Zhong}, \bibinfo{person}{Wenjin Wang}, {and} \bibinfo{person}{Yu Sun}.} \bibinfo{year}{2020}\natexlab{}.
\newblock \showarticletitle{Masked label prediction: Unified message passing model for semi-supervised classification}.
\newblock \bibinfo{journal}{\emph{arXiv preprint arXiv:2009.03509}} (\bibinfo{year}{2020}).
\newblock


\bibitem[Simeunovic et~al\mbox{.}(2021)]%
        {DBLP:journals/corr/abs-2107-13875}
\bibfield{author}{\bibinfo{person}{Jelena Simeunovic}, \bibinfo{person}{Baptiste Schubnel}, \bibinfo{person}{Pierre{-}Jean Alet}, {and} \bibinfo{person}{Rafael~E. Carrillo}.} \bibinfo{year}{2021}\natexlab{}.
\newblock \showarticletitle{Spatio-temporal graph neural networks for multi-site {PV} power forecasting}.
\newblock \bibinfo{journal}{\emph{CoRR}}  \bibinfo{volume}{abs/2107.13875} (\bibinfo{year}{2021}).
\newblock
\showeprint[arXiv]{2107.13875}
\urldef\tempurl%
\url{https://arxiv.org/abs/2107.13875}
\showURL{%
\tempurl}


\bibitem[Trull et~al\mbox{.}(2022)]%
        {trull2022multiple}
\bibfield{author}{\bibinfo{person}{Oscar Trull}, \bibinfo{person}{J~Carlos Garc{\'\i}a-D{\'\i}az}, {and} \bibinfo{person}{Angel Peir{\'o}-Signes}.} \bibinfo{year}{2022}\natexlab{}.
\newblock \showarticletitle{Multiple seasonal STL decomposition with discrete-interval moving seasonalities}.
\newblock \bibinfo{journal}{\emph{Appl. Math. Comput.}}  \bibinfo{volume}{433} (\bibinfo{year}{2022}), \bibinfo{pages}{127398}.
\newblock


\bibitem[Ulanova et~al\mbox{.}(2015)]%
        {ulanova2015efficient}
\bibfield{author}{\bibinfo{person}{Liudmila Ulanova}, \bibinfo{person}{Tan Yan}, \bibinfo{person}{Haifeng Chen}, \bibinfo{person}{Guofei Jiang}, \bibinfo{person}{Eamonn Keogh}, {and} \bibinfo{person}{Kai Zhang}.} \bibinfo{year}{2015}\natexlab{}.
\newblock \showarticletitle{Efficient long-term degradation profiling in time series for complex physical systems}. In \bibinfo{booktitle}{\emph{Proceedings of the 21th ACM SIGKDD International Conference on Knowledge Discovery and Data Mining}}. \bibinfo{pages}{2167--2176}.
\newblock


\bibitem[Velickovic et~al\mbox{.}(2017)]%
        {velickovic2017graph}
\bibfield{author}{\bibinfo{person}{Petar Velickovic}, \bibinfo{person}{Guillem Cucurull}, \bibinfo{person}{Arantxa Casanova}, \bibinfo{person}{Adriana Romero}, \bibinfo{person}{Pietro Lio}, \bibinfo{person}{Yoshua Bengio}, {et~al\mbox{.}}} \bibinfo{year}{2017}\natexlab{}.
\newblock \showarticletitle{Graph attention networks}.
\newblock \bibinfo{journal}{\emph{stat}} \bibinfo{volume}{1050}, \bibinfo{number}{20} (\bibinfo{year}{2017}), \bibinfo{pages}{10--48550}.
\newblock


\bibitem[Woo et~al\mbox{.}(2022)]%
        {woo2022cost}
\bibfield{author}{\bibinfo{person}{Gerald Woo}, \bibinfo{person}{Chenghao Liu}, \bibinfo{person}{Doyen Sahoo}, \bibinfo{person}{Akshat Kumar}, {and} \bibinfo{person}{Steven Hoi}.} \bibinfo{year}{2022}\natexlab{}.
\newblock \showarticletitle{Co{ST}: Contrastive Learning of Disentangled Seasonal-Trend Representations for Time Series Forecasting}. In \bibinfo{booktitle}{\emph{International Conference on Learning Representations}}.
\newblock


\bibitem[Wu et~al\mbox{.}(2020)]%
        {wu2020comprehensive}
\bibfield{author}{\bibinfo{person}{Zonghan Wu}, \bibinfo{person}{Shirui Pan}, \bibinfo{person}{Fengwen Chen}, \bibinfo{person}{Guodong Long}, \bibinfo{person}{Chengqi Zhang}, {and} \bibinfo{person}{S~Yu Philip}.} \bibinfo{year}{2020}\natexlab{}.
\newblock \showarticletitle{A comprehensive survey on graph neural networks}.
\newblock \bibinfo{journal}{\emph{IEEE transactions on neural networks and learning systems}} \bibinfo{volume}{32}, \bibinfo{number}{1} (\bibinfo{year}{2020}), \bibinfo{pages}{4--24}.
\newblock


\bibitem[Yu et~al\mbox{.}(2018)]%
        {Yu2018STGCN}
\bibfield{author}{\bibinfo{person}{Bing Yu}, \bibinfo{person}{Haoteng Yin}, {and} \bibinfo{person}{Zhanxing Zhu}.} \bibinfo{year}{2018}\natexlab{}.
\newblock \showarticletitle{Spatio-Temporal Graph Convolutional Networks: A Deep Learning Framework for Traffic Forecasting}. In \bibinfo{booktitle}{\emph{Proceedings of the 27th International Joint Conference on Artificial Intelligence}}. \bibinfo{pages}{3634–3640}.
\newblock


\bibitem[Zheng et~al\mbox{.}(2020a)]%
        {zheng2020gman}
\bibfield{author}{\bibinfo{person}{Chuanpan Zheng}, \bibinfo{person}{Xiaoliang Fan}, \bibinfo{person}{Cheng Wang}, {and} \bibinfo{person}{Jianzhong Qi}.} \bibinfo{year}{2020}\natexlab{a}.
\newblock \showarticletitle{Gman: A graph multi-attention network for traffic prediction}. In \bibinfo{booktitle}{\emph{Proceedings of the AAAI Conference on Artificial Intelligence}}, Vol.~\bibinfo{volume}{34}. \bibinfo{pages}{1234--1241}.
\newblock


\bibitem[Zheng et~al\mbox{.}(2020b)]%
        {zheng2020distdgl}
\bibfield{author}{\bibinfo{person}{Da Zheng}, \bibinfo{person}{Chao Ma}, \bibinfo{person}{Minjie Wang}, \bibinfo{person}{Jinjing Zhou}, \bibinfo{person}{Qidong Su}, \bibinfo{person}{Xiang Song}, \bibinfo{person}{Quan Gan}, \bibinfo{person}{Zheng Zhang}, {and} \bibinfo{person}{George Karypis}.} \bibinfo{year}{2020}\natexlab{b}.
\newblock \showarticletitle{Distdgl: distributed graph neural network training for billion-scale graphs}. In \bibinfo{booktitle}{\emph{2020 IEEE/ACM 10th Workshop on Irregular Applications: Architectures and Algorithms (IA3)}}. \bibinfo{publisher}{IEEE}, \bibinfo{pages}{36--44}.
\newblock


\end{thebibliography}
\end{document}